\documentclass[a4paper,fleqn]{cas-dc}

\usepackage{graphicx}
\usepackage{caption}
\usepackage{subcaption}
\usepackage{svg}
\usepackage{booktabs} 
\usepackage{longtable}
\usepackage{multirow}
\usepackage{enumitem}
\usepackage{diagbox}
\usepackage{adjustbox}
\usepackage{tcolorbox}
\usepackage{soul, color, xcolor}
\soulregister{\cite}7 
\soulregister{\citep}7 
\soulregister{\citet}7 
\soulregister{\ref}7 
\soulregister{\pageref}7 
\soulregister{\pageref}7 
\sethlcolor{yellow}
\usepackage[ruled,linesnumbered]{algorithm2e}

\usepackage[square, comma, sort&compress, numbers]{natbib}

\usepackage{orcidlink}

\def\tsc#1{\csdef{#1}{\textsc{\lowercase{#1}}\xspace}}
\tsc{WGM}
\tsc{QE}

\begin{document}
\let\WriteBookmarks\relax
\def\floatpagepagefraction{1}
\def\textpagefraction{.001}
\let\printorcid\relax 

\shorttitle{Argument-Centric Causal Intervention Method for Mitigating Bias in Cross-Document Event Coreference Resolution}    

\shortauthors{Yao et al.}

\title[mode = title]{Argument-Centric Causal Intervention Method for Mitigating Bias in Cross-Document Event Coreference Resolution}  

\tnotemark[1]

\tnotetext[1]{This work is a research achievement supported by the "Tianshan Talent" Research Project of Xinjiang (No. 2022TSYCLJ0037), the National Natural Science Foundation of China (No. 62262065), the Science and Technology Program of Xinjiang (No. 2022B01008), the National Key R\&D Program of China Major Project (No. 2022ZD0115800), the National Science Foundation of China (No. 62341206), and the National Science Foundation of China (No. 62476233).}
\author[label1, label2]{Long Yao}  
\ead{107552301321@stu.xju.edu.cn}

\author[label1, label2]{Wenzhong Yang}
\ead{yangwenzhong@xju.edu.cn}\cormark[1]

\author[label1, label2]{Yabo Yin}
\ead{yinyabo@xju.edu.cn}\cormark[1]

\author[label1, label2]{Fuyuan Wei}
\ead{wfy@stu.xju.edu.cn} 

\author[label1, label2]{Hongzhen Lv}
\ead{107552304103@stu.xju.edu.cn}

\author[label1, label2]{Jiaren Peng}
\ead{107552201362@stu.xju.edu.cn}

\author[label1, label2]{Liejun Wang}
\ead{wljxju@xju.edu.cn}

\author[label3]{Xiaoming Tao} 
\ead{taoxm@tsinghua.edu.cn}

\address[label1]{School of Computer Science and Technology, Xinjiang University, Urumqi, 830046, Xinjiang, China}
\address[label2]{Xinjiang Key Laboratory of Multilingual Information Technology, Xinjiang University, Urumqi, 830046, Xinjiang, China}
\address[label3]{Department of Electronic Engineering, Tsinghua University, Beijing, 100084, China}
\cortext[1]{Corresponding author}
\begin{sloppypar}
\begin{abstract}
Cross-document Event Coreference Resolution (CD-ECR) is a fundamental task in natural language processing (NLP) that seeks to determine whether event mentions across multiple documents refer to the same real-world occurrence. However, current CD-ECR approaches predominantly rely on trigger features within input mention pairs, which induce spurious correlations between surface-level lexical features and coreference relationships, impairing the overall performance of the models. To address this issue, we propose a novel cross-document event coreference resolution method based on \textbf{A}rgument-\textbf{C}entric \textbf{C}ausal \textbf{I}ntervention (\textbf{ACCI}). Specifically, we construct a structural causal graph to uncover confounding dependencies between lexical triggers and coreference labels, and introduce backdoor-adjusted interventions to isolate the true causal effect of argument semantics. To further mitigate spurious correlations, ACCI integrates a counterfactual reasoning module that quantifies the causal influence of trigger word perturbations, and an argument-aware enhancement module to promote greater sensitivity to semantically grounded information. In contrast to prior methods that depend on costly data augmentation or heuristic-based filtering, ACCI enables effective debiasing in a unified end-to-end framework without altering the underlying training procedure. Extensive experiments demonstrate that ACCI achieves CoNLL F1 of 88.4\% on ECB+ and 85.2\% on GVC, achieving state-of-the-art performance. The implementation and materials are available at \url{https://github.com/era211/ACCI}.
\end{abstract}


\maketitle

\section{Introduction}
\label{introduction}

\begin{figure}
    \centering
    \includegraphics[width=0.9\columnwidth]{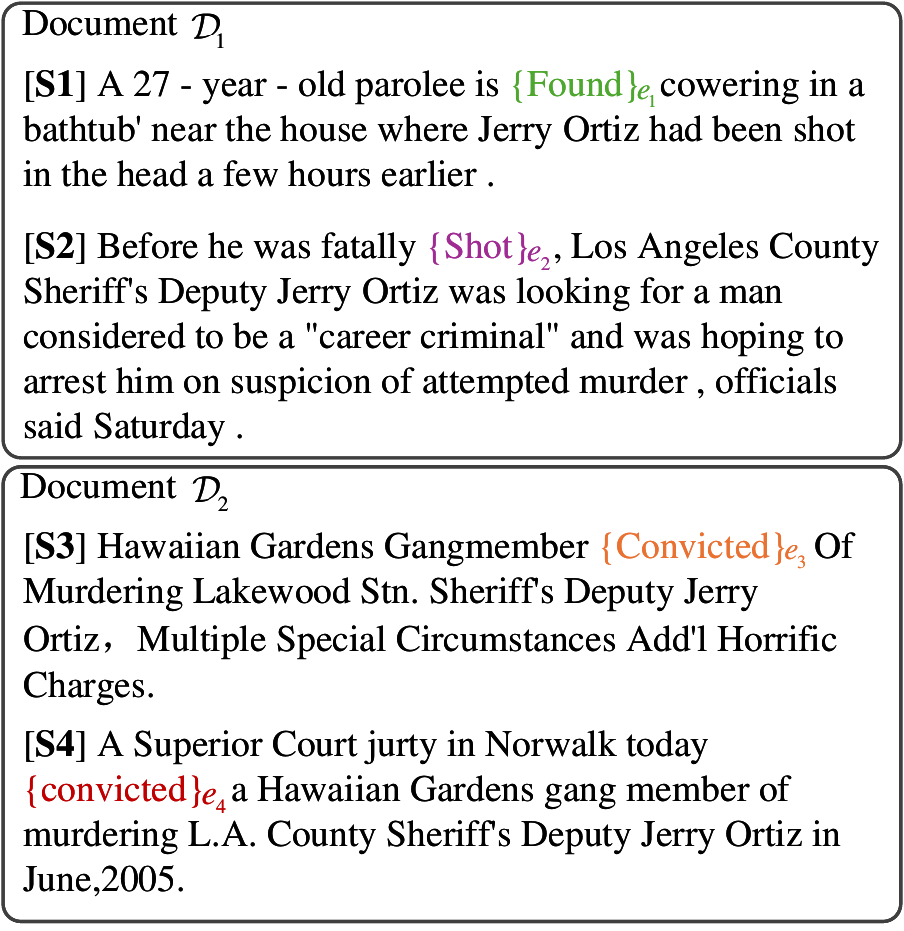}
    \caption{Examples of Event Mentions in Within-Document and Cross-Document Scenarios under the ECR Task. Document $\mathcal{D}_1$ contains mentions $e_1$ and $e_2$, while document $\mathcal{D}_2$ includes mentions $e_3$ and $e_4$. In WD-ECR, the model evaluates intra-document pairs like $(e_3,e_4)$. In CDECR, it analyzes cross-document pairs like $(e_1,e_4)$ to cluster coreferential events across texts.}
    \label{fig:1}
\end{figure}

Event coreference resolution (ECR) is a fundamental task in deep semantic understanding, aiming to identify distinct textual expressions that refer to the same real-world events. It plays a pivotal role in the interpretation and organization of large-scale textual data, with broad applications in knowledge graph construction \cite{postma2018semeval}, information extraction \cite{humphreys1997event}, and question answering \cite{yang2018hotpotqa}.  Fig. \ref{fig:1} illustrates an example of ECR involving two documents $\mathcal{D}_1$ and $\mathcal{D}_2$. Each contains four event mentions: \textit{$e_1$}, \textit{$e_2$}, \textit{$e_3$}, and \textit{$e_4$}. These mentions are paired (e.g. (\textit{$e_1$}, \textit{$e_4$}), (\textit{$e_3$}, \textit{$e_4$})) to determine whether they refer to the same event. Event mentions predicted as coreferent are grouped into a single event chain through clustering. The ECR task is categorized into within-document (WD-ECR) and cross-document (CD-ECR) scenarios, depending on whether the event mentions originate from the same document. WD-ECR addresses mention pairs within a single document (e.g. (\textit{$e_3$}, \textit{$e_4$})), while CD-ECR focuses on mention pairs spanning multiple documents (e.g. (\textit{$e_1$}, \textit{$e_4$})). This work primarily investigates event coreference resolution in the cross-document setting.

\begin{figure}
    \centering
    \includegraphics[width=0.9\columnwidth]{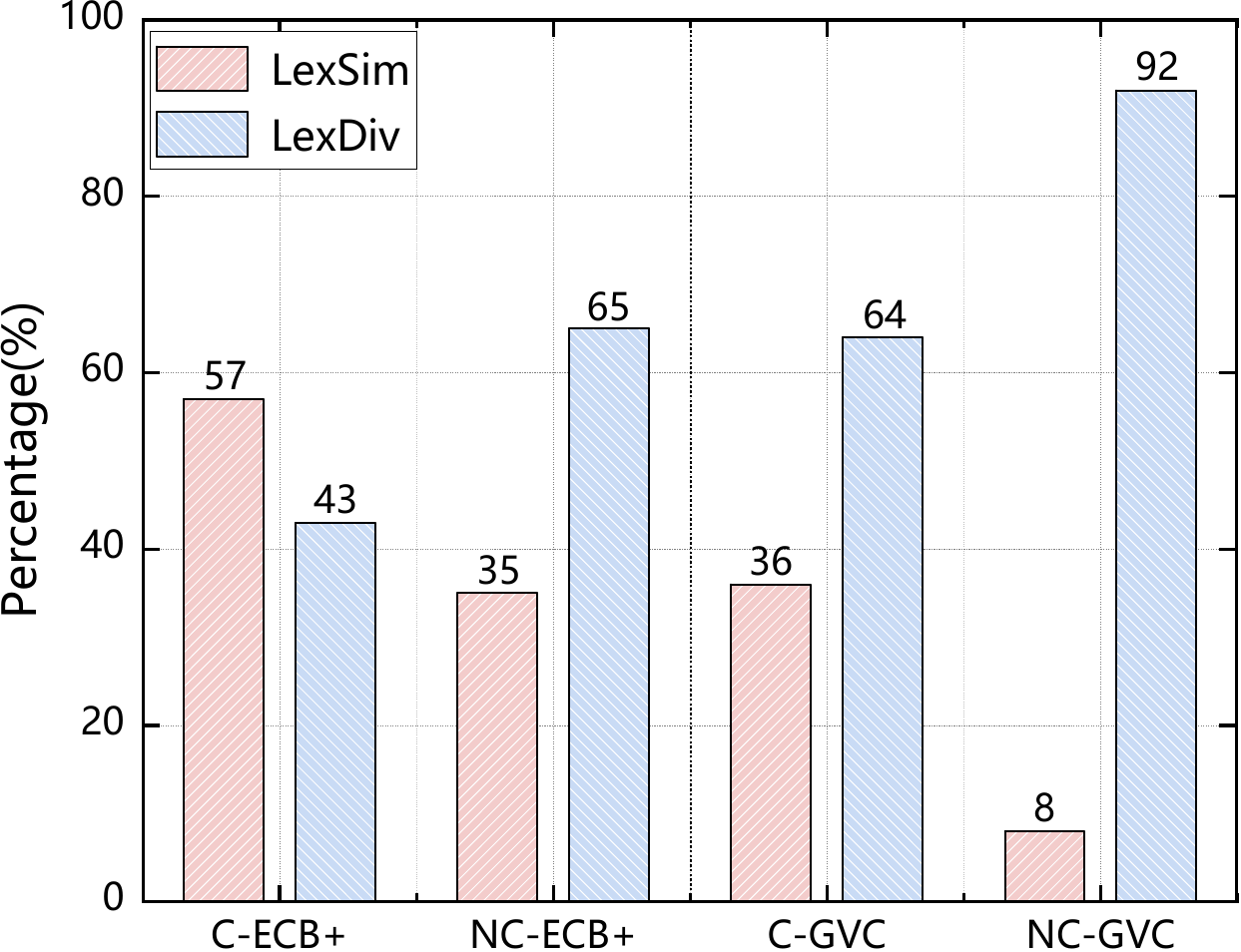}
    \caption{Distribution of “trigger word matching” between mention pairs in the ECB+ and GVC training sets (The left is ECB+ and the right is GVC), with $C$ and $NC$ denoting coreferential and non-coreferential pairs, respectively. $LexSim$ versus $LexDiv$ indicating lexically similar or dissimilar trigger words.}
    \label{fig:2}
\end{figure}

Most existing approaches to the CD-ECR task formulate it as a similarity assessment between event mentions \cite{barhom2019revisiting, ahmed20232}. A commonly adopted paradigm is pairwise representation learning \cite{Yu0R22}, in which pre-trained language models jointly encode event mentions and their surrounding contexts. From these contextualized embeddings, salient features are extracted to train a coreference scorer that estimates the likelihood of two mentions referring to the same event. However, such methods primarily rely on the lexical properties of trigger words and their correlation with coreference labels. Previous study \cite{ding2024rationale} have shown that such scorers often exhibit a strong bias towards superficial lexical matching of trigger words, at the expense of deeper understanding of event arguments. As illustrated in Fig. \ref{fig:2}, coreferential mention pairs tend to share lexically similar triggers, while non-coreferential pairs usually do not. This distributional asymmetry can induce spurious correlations between trigger word similarity and coreference labels, thereby undermining the model's accuracy.

\begin{figure}
    \centering
    \includegraphics[width=1\columnwidth]{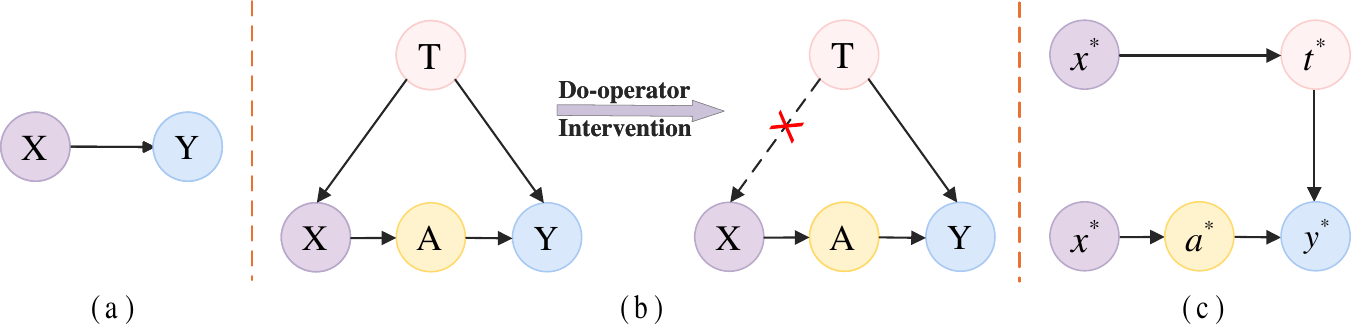}
    \caption{Structural Causal Graph for Event Coreference Resolution. This figure illustrates causal relationships in ECR using a Structural Causal Model. Key variables include event mention pairs ($X$), coreference labels ($Y$), and trigger word matching ($T$) as a confounder. Path $X \leftarrow T \rightarrow Y$ shows how trigger words simultaneously influence mention representation and coreference judgment, creating spurious correlations. Subfigures show: ($a$) baseline causal structure, ($b$) confounding bias from T, and ($c$) Results after controlling for confounding factors.}
    \label{fig:3}
\end{figure}

To elucidate the underlying causes of spurious correlations in CD-ECR, we introduce a structural causal model (SCM) to formally characterize the latent causal mechanisms inherent in this task. As shown in Fig. \ref{fig:3}, the causal dependencies among event pairs $X$, coreference labels $Y$, and confounding factors $T$, are represented through directed edges in a causal graph. Previous studies \cite{ahmed20232, ding2024rationale, ravi2023happens} have underscored that current models tend to overly depend on the lexical matching of trigger words, and this characteristic is modeled as a latent confounder within our causal framework. Specifically, as illustrated in Fig. \ref{fig:3}(b), in addition to the main causal path $X \rightarrow Y$, there exists a confounding path $X \leftarrow T \rightarrow Y$, induced by the confounder $T$. This pathway introduces spurious correlations that obscure the true causal effect of $X$ on $Y$, leading to biased coreference predictions. To effectively mitigate the impact of such spurious associations, we introduce a causal intervention mechanism to neutralize confounding effects.

Inspired by causal intervention \cite{pearl2016causal}, we propose a cross-document event coreference resolution method based on an \textbf{A}rgument-\textbf{C}entric \textbf{C}ausal \textbf{I}ntervention(ACCI). This approach guide the model in distinguishing between the genuine contextual effects of event arguments and the spurious influence of trigger words, thereby promoting unbiased reasoning. Specifically, we begin by modeling the decision-making process of ECR systems using a structured causal graph. As illustrated in Fig. \ref{fig:3}(b), this graph captures the dependencies among event arguments, potential confounders, and other key components of event mentions. Within this framework, we apply the backdoor criterion to design an effective intervention pathway, using the do-operator to isolate the causal effect of confounding variables on the target variable $Y$, thereby blocking confounding influences. This approach provides unbiased estimates of causal effects for the ECR task, as illustrated in Fig. \ref{fig:3}(c). To further quantify the causal contributions of individual factors to model predictions. ACCI generates a series of counterfactual scenarios for each factual instance and compares them to their original counterparts, thereby disentangling the influence of confounding variables and highlighting the causal impact of event arguments in coreference decisions. Notably, the integration of counterfactual reasoning does not modify the model training procedure, which allows it to naturally preserve the inductive biases present in the original data distribution. This design enables the analysis of model behavior under varying semantic conditions without interfering with the optimization objective. Through the joint operation of the Counterfactual Reasoning Module and the Argument-Aware Enhancement Module, ACCI effectively captures the semantic disparities between factual and counterfactual representations, thereby mitigating the model's overreliance on trigger words while enhancing its capacity to extract causal relationships from event arguments.

Specifically, our contributions are as follows:

\begin{enumerate}
    \item We are the first to introduce causal intervention and counterfactual reasoning into the CD-ECR task. The proposed ACCI framework is designed to mitigate spurious correlations in the representation of event mention pairs, without necessitating modifications to the original training paradigm.
    \item We apply the back-door criterion to causally disentangle event constituents from confounding factors. By quantifying the semantic divergence between factual and counterfactual representations. It mitigates spurious correlations introduced by superficial trigger word matching and enhances the model's ability to capture the causal semantics of event arguments.
    \item Extensive experimental results show that ACCI outperforms existing methods, achieving state-of-the-art performance.
\end{enumerate}

\section{Related Work}
\label{sec:Related Work}
\subsection{Event Coreference Resolution}
\label{sec: ECR}
The earliest research on coreference resolution focused on entity coreference resolution, which determines whether different pronouns or noun phrases in the text refer to the same entity. With the growing need to understand event relationships, Bagga et al. \cite{bagga1999cross} first proposed event coreference resolution, and the research focus gradually expanded from entity coreference to event coreference. Compared to entity coreference resolution, event coreference resolution is more complex because events typically involve multidimensional information beyond trigger words, such as time, location, and participants. Early methods for event coreference resolution primarily employed traditional machine learning approaches based on probabilistic models. For example, Lu et al. \cite{lu2017joint} proposed a joint inference model using Markov logic networks (MLNs), which enhanced MLN distributions with low-dimensional unit clauses and implicitly encoded rich features into the model. However, traditional machine learning methods often extract event features containing substantial noise, requiring extensive feature engineering for training. To address the need for integrating diverse event information sources, event coreference resolution is categorized into within-document (WD-ECR) and cross-document (CD-ECR) scenarios, depending on whether events originate from the same document.

\textbf{Within-Document Event Coreference Resolution.} Fang et al. \cite{fang2018employing} leveraged attention mechanisms to extract useful event information, proposing a neural network model with attention (MDAN) that incorporates document-level global reasoning to analyze coreferential event chains. Xu et al. \cite{xu2023corefprompt} designed the task-specific prompts to reframe ECR as a masked language prediction task, enabling concurrent event modeling and coreference resolution within a shared contextual framework. While these methods achieved significant performance over previous state-of-the-art approaches, they underperform in cross-document event coreference resolution tasks due to their inability to handle inter-document event heterogeneity. Yao et al. \cite{yao2023learning} proposed a novel model for within-document event coreference resolution, which addresses the challenges of event coreference by learning and integrating multiple representations of both individual events and event pairs. The model incorporates a variety of linguistically inspired features for individual events, as well as diverse similarity metrics to effectively capture distinctions between event pairs.

\textbf{Cross-Document Event Coreference Resolution.} In recent years, deep learning-based approaches have driven notable progress in CD-ECR. Caciularu et al. \cite{caciularu2021cdlm} proposed CDLM, an extension of Longformer that combines global attention for masked label prediction with local attention elsewhere. Some studies have explored end-to-end approaches. Cattan et al. \cite{cattan2021cross} extended a within-document coreference model to an end-to-end framework that resolves cross-document coreference directly from raw content. Lu et al. \cite{lu2022end} proposed an end-to-end method for predicting event chains directly from raw text. 

Some studies have also adopted pairwise representations for coreference prediction. Barhom et al. \cite{barhom2019revisiting} proposed a model for joint cross-document entity and event coreference, using a learned pairwise scorer for clustering. Event and entity mention representations are enriched via predicate-argument structures to incorporate related coreference clusters. Held et al. \cite{held2021focus} proposed a two-stage approach for cross-document entity and event coreference, inspired by discourse coherence theory. It models attentional states as mention neighborhoods for candidate retrieval, then samples hard negatives to train a fine-grained classifier using local discourse features. Yu et al. \cite{Yu0R22} proposed PAIRWISERL, a pairwise representation learning method that jointly encodes sentence pairs containing event mentions, enhancing the model’s ability to capture structured relationships between events and their arguments. Ahmed et al. \cite{ahmed20232} decomposed ECR into two subtasks to address computational challenges and data distribution skew. The method filters non-coreferent pairs using a heuristic, then trains a discriminator on the balanced dataset for pairwise scoring and coreference prediction. Ding et al. \cite{ding2024rationale} introduced LLM-RCDA, a rationale-centric counterfactual data augmentation method that intervenes on triggers and context, reducing reliance on surface trigger word matching and emphasizing causal associations between events. Graph-based methods have also been applied to the CD-ECR task. Chen et al. \cite{chen2023cross} incorporated discourse structure as global context into CD-ECR to better capture long-distance event interactions. They modeled documents using rhetorical structure trees and represented event mention interactions via shortest dependency paths and lowest common ancestors. Gao et al. \cite{gao2024enhancing} constructed document-level RST trees and cross-document lexical chains to model structural and semantic information, using Graph Attention Networks to learn event representations. This approach improves the modeling of long-distance dependencies in CD-ECR. Chen et al. \cite{chen2025improving} introduced the ECD-CoE task to generate coherent cross-document text for event mentions, addressing limitations in capturing long-distance dependencies and contextual coherence. They encoded the generated text using rhetorical relation trees to extract global interaction features for coreference resolution. Zhao et al. \cite{zhao2025hypergraph} proposed HGCN-ECR, a hypergraph-based method for CD-ECR. It uses BiLSTM-CRF for semantic role labeling, constructs a multi-document event hypergraph centered on triggers, and employs hypergraph convolution and multi-head attention to capture higher-order semantic relations. Ahmed et al. \cite{ahmed2024linear} introduced X-AMR, a graphical event representation, and applied a multi-hop coreference algorithm to linearize ECR over event graphs. Apart from the aforementioned approaches, Nath et al. \cite{nath2024okay} used Free-Text Rationales (FTRs) from LLMs as distant supervision for smaller student models. Through Rationale-Oriented Event Clustering (ROEC) and Knowledge Distillation, their method significantly improved performance on the ECB+ and GVC corpora.

\subsection{Causal Inference}
\label{CI}
Causal inference \cite{pearl2016causal}, as a potential statistical theory originating from statistics and data science, aims to determine whether one variable causes another through statistical and experimental methods. Its core lies in clarifying causal relationships between variables rather than superficial correlations. Causal inference primarily encompasses structural causal graph \cite{neuberg2003causality}, causal intervention \cite{lin2022causal}, \cite{yang2021causal}, and counterfactual reasoning \cite{qian2021counterfactual}, \cite{tang2020unbiased}, which can scientifically identify causal relationships between variables and eliminate potential biases in data. Causal intervention \cite{pearl2009causal} focuses on altering the natural tendency of independent variables to change with other variables to eliminate adverse effects. Counterfactual reasoning \cite{pearl2009causality} describes the imagined outcomes of factual variables in constructed counterfactual worlds.

In recent years, causal inference has been increasingly applied to various down-stream tasks such as recommendation systems \cite{zhang2021causal}, \cite{wang2021clicks}, natural language inference \cite{zhang2021biasing}, \cite{huang2020counterfactually}, computer vision \cite{yang2021deconfounded}, \cite{yang2023context}, and multimodal modeling \cite{zhang2024if}, \cite{chen2023causal}. Mu et al. \cite{mu2023enhancing} employed counterfactual reasoning in event causality identification to estimate the impact of contextual keywords and event pairs on model predictions during training, thereby eliminating inference biases. Chen et al. \cite{chen2023causal} proposed a multimodal fake news debiasing framework based on causal intervention and counterfactual reasoning from a causal perspective. They first eliminated spurious correlations between textual features and news labels through causal intervention, then imagined a counterfactual world to apply counterfactual reasoning for estimating the direct effects of image features.

\subsection{Differences from Existing Methods}
\label{DEM}
Although existing research has achieved significant progress in improving cross-document event coreference resolution (CD-ECR) performance, discussions on mitigating spurious correlations caused by the over-reliance on trigger words in current ECR systems remain insufficiently explored. While Ding et al. \cite{ding2024rationale} alleviated such biases at the data level through counterfactual data augmentation, this approach inevitably incurs additional data annotation costs. Current research trends favor leveraging causal inference to eliminate spurious correlations a strategy that avoids the need for extra labeled data compared to data augmentation methods. In our work, we adopt a causal perspective and employ counterfactual reasoning to simultaneously suppress the model’s over-reliance on trigger words and enhance its perception of causal logic chains among event arguments. Our study primarily focuses on addressing implicit spurious correlations in pairwise event mention representations under cross-document scenarios.
\section{Preliminaries}
\label{sec:Preliminaries}
This section first introduces some basic concepts related to event coreference resolution, followed by an introduction to the fundamental concepts related to causal inference.

\subsection{Key Concepts in Event Coreference Resolutionn}

\begin{enumerate}[label=(\arabic*)]
  \item \textbf{Event}. An event refers to a dynamic or static situation described in text, encompassing actions, state changes, incidents, organizational activities, or other types of activities/phenomena. It typically consists of a trigger word and event arguments, with attributes such as time, location, and participants.
  \item \textbf{Trigger}. The trigger is the core word (often a verb or noun) in the text that explicitly signals the occurrence of an event. The presence of a trigger marks the existence of an event and helps identify its category.
  \item \textbf{Event Mention}. An event mention is a textual expression of a specific event, usually composed of a trigger word and its associated arguments. The same event may be mentioned differently across texts.
  \item \textbf{Event Mention Pair}. In event coreference resolution, determining whether two event mentions refer to the same event is the core task. An event mention pair consists of two event mentions evaluated for coreference.
  \item \textbf{Argument}. An argument in event coreference refers to the entities or participants involved in the event. Common types of arguments include the agent (the doer of the event), the patient (the object affected by the event), as well as time, location, and other contextual factors. Event arguments play a vital role in differentiating events that share similar triggers but differ in contextual semantics or involved participants.
\end{enumerate}

\subsection{Fundamental Concepts in Causal Inference}
\subsubsection{Causal Graph}
\begin{figure}
    \centering
    \includegraphics[width=0.35\columnwidth]{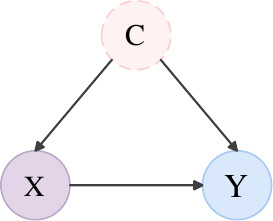}
    \caption{Illustrative of a causal graph, where $X$, $Y$, and $C$ represent the cause, outcome, and mediator, respectively.}
    \label{fig:4}
\end{figure}

In causal inference, the causal graph is widely recognized as a structured tool utilized for representing causal relationships among variables \cite{pearl2016causal}. Formally, a causal graph is formally defined as a directed acyclic graph (DAG), denoted $\mathcal{G}=\{\mathcal{N}, \xi\}$, which consists of a set of variable nodes $\mathcal{N}$ and a set of edges $\xi$ that represent the causal links between these variables. Within a causal graph, several common causal structures are frequently observed, including chains, colliders, and forks. Fig. \ref{fig:4} provides an illustrative example of a causal graph exhibiting a fork structure. The edge $X \rightarrow Y$ specifically indicates that $X$ exerts a causal influence on $Y$. Furthermore, the edge configuration $X \leftarrow C \rightarrow Y$ signifies that $C$ functions as a common cause for both $X$ and $Y$, which represents a typical confounding structure.

\subsubsection{Casual Intervention}
Causal intervention is aimed at evaluating the changes that occur in other variables within a system after a specific variable has been actively controlled or subjected to external manipulation. In contrast to observational analysis, which relies merely on statistical correlations observed under natural conditions, interventional analysis seeks to answer a fundamentally decision-oriented question: \textit{What would happen to the system if the value of a variable were forcibly set?} In traditional statistical modeling, observational data are commonly used to estimate correlations between variables by computing conditional probabilities such as $P(Y|X)$. However, it is crucial to recognize that such conditional probabilities reflect mere associations and don't imply causal relationships. Causal intervention is formalized through the introduction of the $\text{do}(\cdot)$ operator, which explicitly denotes an external manipulation imposed on the system. For instance, $P(Y|\text{do}(X=x))$ denotes the distribution of $Y$ under the condition where variable $X$ is externally set to $x$. This operation explicitly breaks the natural associations that exist along the causal paths within the original system. In the causal graph, the intervention operation can be effectively modeled by removing all incoming edges to the intervened variable, thereby disrupting its original generative mechanism and overriding its value. More precisely, applying $\text{do}(X = x)$ to a variable $X$ in a causal graph $\mathcal{G}$ results in a modified graph $\mathcal{G}_X$, from which the post-intervention distribution of $Y$ can then be evaluated. To accurately estimate such interventional effects, the back-door criterion \cite{pearl2016causal} offers a principled method for identifying a set of confounding variables, enabling the use of Bayes’ theorem \cite{joyce2003bayes} to compute the causal distribution:
\begin{equation}
    \begin{split}
        P\left(Y\middle| do\left(X\right)\right) &= \sum_{c}{P\left(Y\middle| X,c\right)P\left(c\middle| do\left(X\right)\right)} \\
        &= \sum_{c}{P\left(Y\middle| X,c\right)P\left(c\right)}
    \end{split}
    \label{eq:equation1}
\end{equation}
where $c$ denotes specific values taken by the confounding variable $C$.

\subsubsection{Counterfactual Reasoning}
\begin{figure}
    \centering
    \includegraphics[width=0.8\columnwidth]{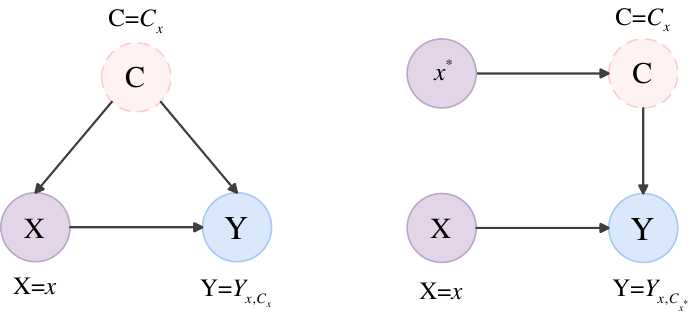}
    \caption{An example of counterfactual reasoning using a causal graph, and $\ast$ denote reference values.}
    \label{fig:5}
\end{figure}

Counterfactual reasoning \cite{pearl2009causality} represents an advanced analytical paradigm within the domain of causal inference. Given factual observations, it investigates how the outcome variable would have responded if a key causal factor had been deliberately set to an alternative value. For example, given a factual assignment $X=x$ yields the outcome $Y=y$, a counterfactual query seeks to answer: \textit{what would the value of $Y$ have been if $X$ had been set to another value $x^{'}$?} In contrast to traditional statistical inference, counterfactual reasoning seeks to estimate potential outcomes under alternative hypothetical scenarios. It enables models to distinguish genuine causal effects from spurious correlations that arise from surface-level statistical patterns \cite{pearl2022probabilities}.

As illustrated in Fig. \ref{fig:5}, in a causal graph exhibiting a fork structure, where
$X=x$ and a confounding variable $C$ exists, the outcome variable $Y$ can be denoted as:

\begin{equation}
    Y_{x,C_x}=Y\left(X=x,C=C_{(X=x)}\right)
    \label{eq:equation2}
\end{equation}

Under the counterfactual scenario, where the value of $x$ is altered to $x^\ast$, the corresponding outcome $Y$ can be represented as:
\begin{equation}
    Y_{x^\ast,C_{x^\ast}}=Y\left(X=x^\ast,C=C_{(X=x^\ast)}\right)\ 
    \label{eq:equation3}
\end{equation}

\section{Methodology}
\label{sec:Methodology}

\begin{figure*}
    \centering
    \includegraphics[width=1.9\columnwidth]{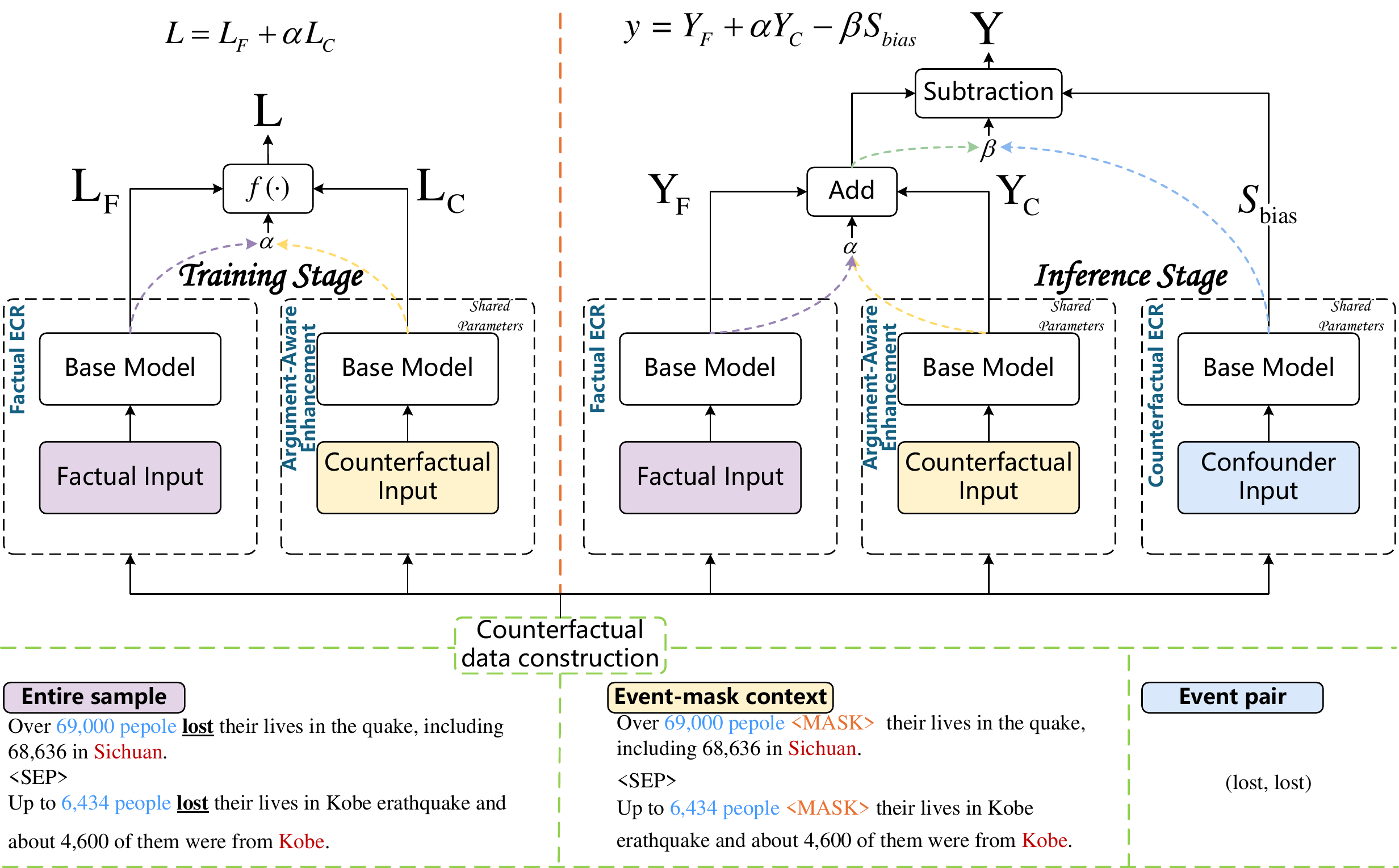}
    \caption{The overall architecture of the ACCI framework.}
    \label{fig:6}
\end{figure*}

In this paper, we introduce an argument-centric causal intervention framework, named ACCI, to mitigate bias in cross-document event coreference resolution. Built upon two widely adopted backbone architectures \cite{ahmed20232, held2021focus}, our approach integrates causal intervention with counterfactual reasoning to effectively alleviate spurious correlations arising from lexical matching of trigger words. Simultaneously, we enhance the model's perception capabilities regarding event arguments based on counterfactual reasoning principles. During the training phase, this approach calibrates model biases by contrasting the semantic differences between factual and counterfactual features. And it strengthens the alignment capabilities among events mention arguments. It significantly improves the model's accuracy in complex semantically ambiguous scenarios. The architecture of the proposed model is illustrated in Fig. \ref{fig:6}.

\subsection{Task Definition}
\label{TD}
Given a document set $\mathcal{D}=\{{d}_1,{d}_2,\dots,d_{m}\}$, where $m=|\mathcal{D}|$. Each document ${d}_i$ contains multiple event mentions. The set of all event mentions in $\mathcal{D}$ constitutes the event mention set $E = \{e_1, e_2, \ldots, e_{n}\}$. For candidate event mention pairs $(e_i,e_j)$ from the same subtopic, the goal of the ECR is to learn a mapping function $\mathscr{F}$ to predict the coreference relation between these two event mentions: $\mathscr{F}:(e_i,e_j) \rightarrow {y}$, where $y\in Y=\{\Upsilon,\Theta\}$. Here, $y=\Upsilon$ indicates that $e_i$ and $e_j$  refer to the same events, and $y=\Theta$ indicates that they refer to different events. Subsequently, event mention pairs predicted to be coreferent are clustered to represent distinct events.

\subsection{Pairwise Event Coreference}
\label{BM}
In this section, we briefly introduce the base model used. The task is generally divided into two primary stages: candidate mention retrieval and coreference resolution.

\textbf{Candidate Mention Retrieval Phase}: To ensure fair comparison with previous work, we employ two candidate event mention pair retrieval strategies: (1) a discourse coherence theory-based approach for generating event mention pairs \cite{held2021focus}; and (2) a heuristic algorithm for the efficient filtering of non-coreferential pairs \cite{ahmed20232}.

\textbf{Coreference Resolution Phase}: To enable fine-grained evaluation of candidate mentions, a pairwise classifier is constructed and trained accordingly. Specifically, for each event mention pair $(\mathcal{A},\mathcal{B})$, the trigger words of both mentions are enclosed with a special marker $($<m>$ \text{and} $</m>$)$ to explicitly signal their roles. Formally, for each event mention, It can be denoted as:
\begin{equation}
    \mathcal{M} =\text{prefix} \ \oplus <m> \oplus \ \mathcal{T}_\mathcal{A} \oplus </m> \oplus \ \text{suffix}
    \label{eq:equation4}
\end{equation}
where $\mathcal{M}$ denotes event mention, the symbol $\oplus$ denotes concatenation, and $\mathcal{T}_\mathcal{A}$ is the trigger word of event $\mathcal{A}$.
Next, the textual content of the marked mentions $\mathcal{A}$ and $\mathcal{B}$ is combined into a joint sequence as input to a model. The input sequence is structured as:
\begin{equation}
    X = [[CLS],\mathcal{M}_{\mathcal{A}},[SEP],\mathcal{M}_{\mathcal{B}},[SEP]]
    \label{eq:equation5}
\end{equation}
where $[CLS]$ and $[SEP]$ are standard special tokens. This sequence is fed into a PLM-based cross-encoder to obtain hidden state representations:
\begin{equation}
    H = Cross\_Encoder(x_1,x_2,\dots,x_i,\dots,x_j)
    \label{eq:equation6}
\end{equation}
where $H\in \mathbb{R}^{L\times d}$, $L$ is the sequence length and $d$ denotes the dimension of the feature representation, with $x_i\in X$ representing different event mention pairs.

Through average pooling operations, we derive event representations $E_\mathcal{A}$ and $E_\mathcal{B}$ centered on trigger words, representing the hidden layer embeddings corresponding to the triggers in event mentions $\mathcal{A}$ and $\mathcal{B}$, respectively. Simultaneously, we extract the global pooled representation $H_{CLS}$ to capture the global semantic context of the entire input. These representations are then concatenated to form a unified event mention pair representation:
\begin{equation}
    h = tanh(W^T(H_{[CLS]}||E_\mathcal{A}||E_\mathcal{B}||E_\mathcal{A}\odot E_\mathcal{B}]))
    \label{eq:equation7}
\end{equation}
where  $W^T$ is a learnable weight matrix, the symbol $||$ denotes concatenation. The symbol $\odot$ denotes element-wise multiplication, which is used to capture fine-grained interactions between event mentions. The concatenated representation is fed into a MLP layer to predict the coreference probability:
\begin{equation}
    P_{ECR} = MLP(W^T_ph)
    \label{eq:equation8}
\end{equation}
where $W^T_p$ is the weight matrix, and $P_{ECR}$ denotes the predicted coreference score. Then, a clustering algorithm groups event mentions predicted as coreferential to form clusters representing coreferent events. As briefly illustrated in Algorithm \ref{alg:alg1}, this process consolidates pairwise predictions into globally coherent event chains.

\begin{algorithm*}
    \SetAlgoLined 
	\caption{Factual ECR Inference Phases}
    \label{alg:alg1}
	\KwIn{ Document set $\mathcal{D}$, event mentions in Document set $\mathcal{M}$, gold/clustered topics $\mathcal{T}$ and sentences of mentions $\mathcal{S}$.} 
    
	\KwOut{Coreferential event clustering $\mathcal{C}$}
        \tcc{Candidate Pair Generation and Scoring}
        pairs $\gets \text{RetrieveCandidatePairs}(\mathcal{M}, \mathcal{T}, \mathcal{S})$
        \\
        Core\_Score $\gets \text{ComputeCoreScores} (pairs,  \mathcal{S})$
        \\
        \tcc{Filter high-confidence pairs based on threshold}
        \If{Core\_Score $\geq$ 0.5}{
            Determine the current pair of event mentions as a coreferent pair:
            \\
            LikelyPairs  $\gets (e_i, e_j) \in$ pairs
            }
        \tcc{Initialize Clusters}
        $C \gets \text{InitializeClusters}(\mathcal{M})$
        \\
        \tcc{Iterative Merging}
        \For{$(e_i, e_j) \in LikelyPairs$}{
            $c_i \gets \text{GetCluster}(\mathcal{C}, e_i)$ \\
            $c_j \gets \text{GetCluster}(\mathcal{C}, e_j)$ \\
            \If{$c_i \neq c_j$}{
                $s_{clust} \gets \text{ComputeClusterScore}(c_i, c_j)$ \\
            \If{$s_{clust} > \tau_{\text{cluster}}$}{
                $C \gets \text{MergeClusters}(\mathcal{C}, c_i, c_j)$ \\
                }
            }
        }
    \KwRet$\mathcal{C}$
\end{algorithm*}

\subsection{Causal Analysis and Intervention of ECR}
\label{CAI}
In this section, based on the structured causal graph constructed, we conduct a comprehensive causal analysis to systematically investigate the inherent bias present in baseline ECR systems. Subsequently, we employ the backdoor adjustment method to implement a causal intervention aimed at blocking the non-causal paths between trigger word matching and event coreference decisions.

\subsubsection{Causal Analysis}
We introduce structural causal models (SCM) \cite{pearl2009causality} to systematically analyze the source of bias in the CD-ECR baseline systems and characterize their underlying decision mechanisms. Specifically, we formalize the model's prediction process as:
\begin{equation}
    Y=\sigma\left(X,U\right)=\sigma\left(X_T,X_A,U\right)
    \label{eq:equation9}
\end{equation}
where $Y$ denotes the model's prediction regarding whether two event mentions are coreferent; $X$ represents the observed input features, comprising trigger matching related features $X_T$ and contextual information pertinent to event semantic coreference $X_A$, such as time, location, participants, and actions; $U$ represents unobservable latent ariables. The operation $\sigma\left(\cdot\right)$  is a mapping function that describes how these explicit and latent factors jointly influence the model's decision process.

Although the CD-ECR system is theoretically expected to rely on rich semantic argument information when identifying coreference relations \cite{cybulska2015bag}, many existing methods exhibit an overreliance on surface-level matching between event triggers. This method introduces bias, encouraging the system to learn spurious correlations between trigger words matching and coreference labels, while neglecting deeper semantic consistency.

To further elucidate the origins of this bias, we model and analyze the model's decision mechanism using the established causal graph.  As illustrated in Fig. \ref{fig:3}(b). We expect the model to follow a causal path from the input features $X$ via the semantic argument information $A$  to the prediction $Y$, represented as $X \rightarrow A \rightarrow Y$, in order to achieve an accurate semantically driven prediction. However, the trigger word matching feature $T$ often influences the input representation $X$. Subsequently, it affects the model's decision, thereby forming a path such as $T \rightarrow X \rightarrow A \rightarrow Y$. This path introduces a non-causal influence on the event coreference prediction $Y$,  causing the model to establish a spurious correlation between trigger word matching and coreference decisions. According to causal theory \cite{pearl2009causal}, the trigger matching feature $T$ effectively acts as a confounder between the input $X$ and the predication $Y$. This confounder introduces bias through the backdoor path $X \leftarrow T \rightarrow Y$, resulting in the model overly relying on superficial semantic similarity during inference while neglecting the modeling of contextual semantic consistency. 


\subsubsection{Causal Intervention Through Backdoor Adjustment}
To mitigate decision bias introduced by trigger words in event coreference resolution, we further incorporate an intervention mechanism grounded in causal analysis. Specifically, we employ the $\text{do}(\cdot)$ operator to simulate the model's true response to input event pairs under conditions where the influence of trigger words is controlled or removed. This process fundamentally aims to disrupt spurious correlations induced by surface-level matching features, thereby enhancing the model's capacity to focus on contextual semantic consistency.

Traditional CD-ECR systems typically leverage likelihood estimation $P(Y|X)$ for predictions. However, this modeling approach inherently fails to distinguish the causal contributions of trigger words versus semantic arguments, leading to bias introduced along the confounding path $X \leftarrow T \rightarrow Y$. As depicted in Fig. \ref{fig:3}(b), we define the following interventional objective:
\begin{equation}
    P\left(Y\middle| do\left(X\right)\right)=\sum_{t\in\tau}{P\left(Y\middle| X,A,T=t\right)P\left(T=t\right)}
    \label{eq:equation11}
\end{equation}

This formula is predicated on the backdoor adjustment principle \cite{chen2023causal}. where the operation $\text{do}(X)$ signifies artificially setting the input features $X$, thereby rendering them independent of the trigger word $T$.  The variable $\tau$ denotes the domain of $T$. In contrast to the prediction distribution $P(Y|X)$ in observational settings, it effectively severs the confounding path emanating from trigger words by replacing the conditional distribution $P(T|X)$ with the marginal distribution $P(T)$.

In the CD-ECR task, given that event pairs are distributed across multiple documents, the consistency of trigger word occurrences often provides an unreliable indication of coreference relationships.  Instead, it can mislead the model to disregard more informative cross-document semantic arguments. Our intervention strategy explicitly diminishes the model's reliance on trigger words through targeted sample feature adjustments, guiding it to learn a more robust judgment mechanism predicated on semantic arguments. To further augment the model's causal explanatory power at the individual sample level and to explore the latent impact of varying trigger word configurations, we further incorporate a counterfactual inference mechanism.

\subsection{Counterfactual ECR}
\label{CECR}
We first introduce the Counterfactual Reasoning Module, which systematically quantifies and mitigates surface bias induced by trigger word matching through the construction of hypothetical scenarios contrary to observed facts. Subsequently, we elucidate the Argument-Aware Enhancement Module, designed to further improve the model's ability to determine the coreference relationship of events without interference from trigger words,  especially by prioritizing the processing of the core semantic information of event arguments.

\subsubsection{Counterfactual Reasoning Module}
Considering that existing CD-ECR models \cite{ahmed20232}, \cite{held2021focus}, and \cite{Yu0R22} frequently exhibit spurious correlations caused by confounding factors,  we pose the following counterfactual query: "\textbf{\textit{What would the coreference resolution outcome be if trigger word information in the event mention pair were unobserved?}}" 
To accurately quantify and remove the spurious correlations induced by trigger word matching, we adopt Pearl’s three-step framework of abduction, intervention, and prediction to simulate such counterfactual scenarios  \cite{pearl2009causality}. This framework enhances the model’s ability to capture the causal effect of event arguments on coreference decisions.  By comparing predictions under factual and counterfactual conditions, it enables precise quantification and mitigation of spurious biases. Specifically, we simulate the effect of trigger word information on coreference decisions by constructing a counterfactual reasoning pathway based on the presence of the event pair $(x_{t_1}, x_{t_2})$.

\textbf{Constructing Counterfactual Event Embeddings}: We first partition the original input sequence $X=\{x_1,\ldots,x_n \}$ into trigger subsequences $X_{trg}=\{x_t \mid t \in T\}$ and context subsequences $X_{arg}=X \backslash X_{trg}$ via masking operations. This allows the model to separately learn spurious correlations from triggers and true causal information from context in decoupled subspaces. Using a cross-encoder, we obtain counterfactual event embeddings $h_{x_t}^{cf}$, which strip away contextual argument information and retain only lexical features of the triggers:
\begin{equation}
    h_{x_t}^{cf}=Cross\_Encoder\left(x_{t_1},x_{t_2}\mid X_{arg}=\emptyset\right),t_i \in T
    \label{eq:equation12}
\end{equation}
where $X_{arg}=\emptyset$ denotes unobserved contextual argument information. What distinguishes counterfactual embeddings from conventional ones is that the former primarily omit contextual semantics and instead encode solely trigger-specific lexical features.

\textbf{Computing Bias-Aware State Representations}: We then compute a joint representation space mapping using:
\begin{equation}
    h_E=tanh\left(W_f^T\left[\Phi_c;h_{x_{t_1}}^{cf};h_{x_{t_2}}^{cf};h_{x_{t_1}}^{cf}\odot h_{x_{t_2}}^{cf}\right]\right)
    \label{eq:equation13}
\end{equation}
where $\Phi_c\in R^d$ is a empty‐context placeholder that simulates zero contextual input, $W_f\in R^{\left(d+2d_{x_t}\right)\times d_h}$ is a trainable weight matrix, and $d_{x_t}$, $d_h$ are the embedding and hidden‐state dimensions, respectively.

\textbf{Quantifying Spurious Correlation}: Finally, we quantify surface-level bias via linear projection to compute pseudo-association strength between event pairs:
\begin{equation}
    Score=MLP\left(W_E^Th_E\right)
    \label{eq:equation14}
\end{equation}
where $W_E\in R^{d_h\times2}$ are the classification weights. The score represents the degree to which spurious correlation arises solely from trigger words.

By applying this approach, we mitigate the spurious signals caused by the lexical similarity of trigger words, allowing the model to concentrate on the genuine causal relationships among events. Thereby it has enhanced the accuracy of cross-document event coreference resolution.

\subsubsection{Argument-Aware Enhancement Module}
After the Counterfactual Reasoning Module successfully quantified and mitigated the spurious correlations introduced by trigger words, we enhanced the ACCI framework's argument awareness by developing an argument-aware enhancement path based on event argument information $X_{arg}$. This path is designed to fully leverage the core semantic information provided by the event arguments.  It enables the model to diminish its overreliance on trigger words, while prioritizing the semantic contribution of argument roles in determining coreference relations, thereby achieving more accurate coreference resolution.

\textbf{Argument-Centric Embedding Extraction}:We input $X_{arg}$ into a cross-encoder to obtain the contextualized embedding of the [CLS] token, denoted as $\mathcal{C}_{arg}$. The resulting embedding aggregates semantic information from all arguments:
\begin{equation}
    \mathcal{C}_{arg}=Cross\_Encoder(X_{arg})
    \label{eq；equation15}
\end{equation}
where $\mathcal{C}_{arg}$ reflects contextual information only, distinct from the $[CLS]$ representation derived from full event pairs.

\textbf{Simulating Argument-Only Scenarios}: To simulate event coreference resolution based solely on argument information, we introduce a learnable constant vector $\Phi \in \mathbb{R}^d$, which replaces the original trigger representation and serves as a placeholder for an empty event. Concatenate $\mathcal{C}_{arg}$ with two instances of $\Phi_E$ (empty trigger placeholders) and apply a nonlinear transformation via shared parameters $W_f$:
\begin{equation}
    h_{arg}=tanh\left(W_f^T\left(\left[\mathcal{C}_{arg};\ \Phi_E;\ \Phi_E\right]\right)\right)
    \label{eq:equation16}
\end{equation}
where $W_f$ denotes a learnable shared parameter, and $\Phi$ represents a placeholder for event representation.

To support more effective learning of semantic representations, we introduce a learnable placeholder vector $\Phi \in \mathbb{R}^d$, which substitutes the original trigger embedding. This design encourages the model to rely primarily on argument-related contextual cues, thereby simulating an argument-only inference scenario. Argument-centric embeddings are concatenated with these placeholders and passed through a non-linear transformation parameterized by a shared weight matrix $W_f$, ensuring consistent representation learning across modules. This shared parameterization ensures consistent representation learning across modules, aligns latent spaces, and prevents divergence within the argument-aware enhancement module while fostering synergistic learning of contextual features. 

\subsection{Training and Debiased Inference}
\label{Debiasing}
We propose a training and debiased inference strategy to enhance the model's causal validity in CD-ECR. This strategy introduces a joint optimization objective during training, which guides the model to prioritize contextual event argument information. Simultaneously, in the inference phase, a causal inference mechanism is employed to explicitly identify and mitigate surface bias stemming from trigger word matching, thereby achieving more reliable coreference decisions. Specifically, during the training phase, we jointly optimize two objectives: one from the factual prediction module, which operates on complete samples, and another from the semantic enhancement module, which focuses on contextual argument information. This joint training encourages the model to pay greater attention to the event arguments embedded in context, thereby enhancing its ability to capture accurate event coreference relationships. Importantly, the loss associated with the counterfactual reasoning module is not involved in the backpropagation process. This design ensures that the module functions solely for bias estimation and does not influence the parameter updates of the other components. The overall training objective is formulated as:
\begin{equation}
    Loss\ =\ Loss_F\ +\ \alpha Loss_C
    \label{eq:equation17}
\end{equation}
where $Loss_F$ denotes the binary cross-entropy loss computed over $P_F$, which represents the model's predictions based on the input and potential spurious signalls. In contrast, $Loss_C$ is the binary cross-entropy loss computed on $P_C$, which encourages the model to make predictions relying solely on contextual argument information, thereby mitigating the lexical bias caused by trigger word interference. The two loss terms are combined using a balancing coefficient $\alpha$. The process is denoted as:
{\small
\begin{equation}
    Loss_{F/C}\ =\ -\frac{1}{m}\sum_{i=1}^{m}\left(y_i\cdot log\ \widehat{y_i}+\left(1-y_i\right)\cdot log\ \left(1-\widehat{y_i}\right)\right)
    \label{eq:equation18}
\end{equation}
}
where $y_i$ and $\widehat{y_i}$ represent the ground-truth and predicted coreference probabilities for the $i$-$th$ input within a training batch.

During the inference phase, a three-step causal debiasing procedure is applied to eliminate spurious correlations induced by trigger-word similarity. First, we mask trigger words within event mentions and construct context-enhanced inputs that explicitly encode arguments such as time, location, and participants. These inputs are encoded by a shared encoder to produce argument-centric semantic representations, from which we derive the argument-enhanced prediction probability $P_{C}$. This encourages the model to base its coreference decisions primarily on argument-level semantic consistency. Second, we construct counterfactual inputs containing only the trigger words. These inputs are passed through the same encoder to generate trigger embeddings, which are then used by the counterfactual reasoning module to estimate the confounding bias term $S_{bias}$. This bias reflects the spurious predictive signal arising from surface-level trigger similarity, corresponding to the confounding path $X\leftarrow T \rightarrow Y$. Finally, the original coreference prediction $P_{F}$ is linearly combined with the auxiliary score $P_C$, and the bias term $S_{bias}$ is subtracted, resulting in the final debiased prediction:
\begin{equation}
    y = P_{F}+\alpha P_C-\beta S_{bias}
    \label{eq:equation19}
\end{equation}
where $\alpha$ denotes the weighting coefficient for the counterfactual prediction probability $P_C$, which reflects the model's dependence on argument-centric causal information. $\beta$ denotes the weight assigned to the bias signal $S_{bias}$, which captures the spurious influence introduced by superficial trigger-level features. This inference pipeline not only integrates the causally informative cues encoded in argument semantics, but also explicitly suppresses misleading correlations, thereby encouraging the model to make coreference decisions grounded in more robust and meaningful signals.


\section{Experiments and Analysis}
\label{sec:Experiments and Analysis}
This section details the datasets, evaluation metrics, baseline models, and experimental setups, followed by comprehensive analysis of results. We further validate the effectiveness of each module through ablation studies and investigate the impact of confounding factors via case studies.

\subsection{Datasets and Evaluation Metrics}
\label{DEM}
\subsubsection{Datasets}
To validate the effectiveness of our proposed method, we conducted experiments on two widely used benchmark datasets: Event Coreference Bank Plus (ECB+) \cite{cybulska2014using} and the Gun Violence Corpus (GVC) \cite{vossen2018don}. These datasets span diverse topics and linguistic contexts, providing a robust basis for evaluating the generalizability of the ACCI-ECR framework.

\textbf{ECB+}: This dataset was introduced by Cybulska and Vossen \cite{cybulska2014using} in 2014 and is currently a widely used standard for cross-document event coreference resolution research. It was created as an extension to the original ECB dataset. The expanded ECB+ dataset enhances the study of cross-document event coreference resolution by drawing on a large collection of news articles that cover various types of events, with coreference relations annotated across multiple documents. The corpus includes text spans marking event mentions and information on how these events are linked via coreference. It comprises 982 articles spanning 43 different topics, with 6,833 event mentions connected by 26,712 coreference links and 8,289 entity mentions connected by 69,050 coreference links. In common usage, documents from topics 1 through 35 are divided into training and development sets (with the development set drawn from topics 2, 5, 12, 18, 21, 34, and 35), and documents from topics 36 through 45 are assigned to the test set.

\textbf{GVC}: Introduced by Vossen et al. \cite{vossen2018don} in 2018, the Gun Violence Corpus is an English dataset specifically designed for event coreference resolution. It comprises texts related to gun-violence incidents collected from multiple sources—news websites, social media platforms, reports, and blogs—all centered on a single theme (gun violence) and exhibiting a wide range of reporting styles, event descriptions, and subsequent reactions. The corpus contains 510 articles and features high lexical variability. In total, GVC includes 7,298 event mentions linked by 29,398 coreference relations.

Following previous work, we train and evaluate our method on the annotated event mentions. Detailed statistics for the ECB+ and GVC datasets are provided in Table~\ref{tab:table1}. For ECB+, we adopt the data split of Cybulska and Vossen \cite{cybulska2015bag}. Meanwhile, we follow the split specified by Bugert et al. \cite{bugert2021generalizing} for GVC dataset.

\begin{table*}[htbp]
  \centering
  \small
  \setlength{\tabcolsep}{18pt} 
  \caption{Statistics of ECB+ and GVC datasets.}
  \begin{tabular}{lllllll}
    \toprule
    & \multicolumn{3}{c}{ECB+} & \multicolumn{3}{c}{GVC} \\
    \cmidrule(r){2-7}
    & Train & Dev & Test & Train & Dev & Test \\
    \midrule
    Topics                    &   25   &   8    &   10   &   1    &   1    &   1    \\
    Documents                 &  574   &  196   &  206   & 358    &  78    &  74    \\
    Event mentions            & 3808   & 1245   & 1780   & 5313   & 977    & 1008   \\
    Event clusters            & 1527   &  409   &  805   & 991    & 228    & 194    \\
    Event singletons          & 1116   &  280   &  623   & 157    &  70    &  43    \\
    Cross-document pairs      &169798  &51849   & 83191  & 60142  & 9193   &10660   \\
    \bottomrule
  \end{tabular}
  \label{tab:table1}
\end{table*}

\subsubsection{Evaluation Metrics}
In this study, consistent with the approach of Ding et al. \cite{ding2024rationale}, and since we did not perform mention detection, we used the $\text{B}^3$ F1 metric proposed by Bagga and Baldwin \cite{bagga1998entity} to select the best model during the training phase. Among various coreference evaluation metrics, $\text{B}^3$ F1 was chosen for its robustness in settings lacking mention detection, as evidenced by the analysis of Moosavi and Strube \cite{moosavi2016coreference}. For comprehensive comparison with existing research, we also report the F1 for $\text{LEA}$ \cite{moosavi2016coreference}, $\text{MUC}$ \cite{vilain1995model}, and $\text{CEAFe}$ \cite{luo2005coreference}, as well as $\text{CoNLL}$ F1, which is the arithmetic mean of the $\text{MUC}$, $\text{B}^3$, and $\text{CEAFe}$ F1.

Specifically, $\text{MUC}$ primarily focuses on the completeness of coreference chains, $\text{B}^3$ emphasizes the correct matching of entities within chains, and $\text{CEAFe}$ enhances adaptability to diverse textual structures by incorporating entity category information. $\text{LEA}$, through its weighting mechanism, highlights the resolution quality of critical entities. Since each metric has strengths and limitations, multiple evaluation methods are typically combined to comprehensively assess system performance. Ultimately, we calculate the F1 of $\text{MUC}$, $\text{B}^3$, and $\text{CEAFe}$, take their average as the overall evaluation benchmark ($\text{CoNLL}$ F1), and report $\text{LEA}$ to reflect the model’s capability in long-distance coreference resolution. Additionally, $\text{MUC}$ measures performance by calculating the minimal number of link adjustments required to align predicted and annotated coreference chains, while $\text{B}^3$ addresses $\text{MUC}$’s limitations by prioritizing individual mention-level matching. $\text{CEAFe}$ evaluates the effectiveness of event coreference resolution through entity information, and $\text{LEA}$ further refines precision for core entities via weighted allocation, revealing the system’s accuracy in resolving critical mentions. This multi-metric approach ensures a balanced and nuanced assessment of coreference resolution systems.

\subsection{Implementation Details}
\label{ID}
Our model is implemented using PyTorch \cite{imambi2021pytorch}. We use AdamW \cite{LoshchilovH19} to optimize the model parameters. To ensure a fair comparison with the main baseline, we followed their experimental setup \cite{ahmed20232}. Specifically, we adopt their heuristic method for data construction. We set the model learning rate to $1e-5$ and the classifier learning rate to $1e-4$. All experiments are conducted in four Nvidia Tesla A40 Gpus.

\subsection{Baseline}
\label{baseline}
To comprehensively evaluate our ACCI model, we conduct comparative experiments with various existing methods on the ECB+ and GVC datasets:

\begin{enumerate}[label=(\arabic*)]
    \item \textbf{Cross-Document Language Modeling (CDLM)}: Caciularu2021 \cite{caciularu2021cdlm} pretrains multiple related documents to build a general-purpose language model for multi-document settings.
    \item \textbf{End-to-End Methods}: Cattan2021 \cite{cattan2021cross} develop an end-to-end model that performs cross-document coreference resolution directly from raw text.
    \item \textbf{Graph-Structured Methods}: Chen2023 \cite{chen2023cross}, Gao2024 \cite{gao2024enhancing}, Chen2025 \cite{chen2025improving}, and Zhao2025 \cite{zhao2025hypergraph} leverage discourse graphs or hypergraph convolutional neural networks combined with structured information across documents to enhance the representation and aggregation of event coreference features.
    \item \textbf{Pairwise Representation Learning}: Barhom2019 \cite{barhom2019revisiting}, Held2021 \cite{held2021focus}, Yu2022 \cite{Yu0R22}, Ahmed2023 \cite{ahmed20232}, and Ding2024 \cite{ding2024rationale} treat event mention pairs as input and employ a cross-encoder framework to learn fine-grained coreference features between mentions.
    \item \textbf{Large language model (LLM)-based Methods}: Nath2024 \cite{nath2024okay} use a large language model as a teacher via rationale-aware clustering and knowledge distillation for cross-document event coreference resolution. We also compare the performance of Llama and GPT-3.5-Turbo on this task. The prompt is designed as follows:

\end{enumerate}

\begin{tcolorbox}[colback=gray!20, left=1mm, right=1mm, top=1mm, bottom=1mm,  width=\linewidth] 
\textbf{Prompt}: You are an expert event‐coreference annotator. You will be given two sentences, each containing exactly one event mention wrapped in <m> and </m> tags. Your task is to decide whether these two marked mentions refer to the same real‐world events or to different events. Please think step by step from multiple perspectives. Your final answer must be exactly one word.\\
\textbf{Input format}:\\
Sentence 1 is: $\{sentence_1\}$\\
Sentence 2 is: $\{sentence_2\}$
\end{tcolorbox}
\vspace{-2mm}


\subsection{Experimental Results}
\label{ER}
\subsubsection{Results on the ECB+ Dataset}
To evaluate the effectiveness of the proposed approach, we implemented ACCI within two representative backbone architectures \cite{ahmed20232},\cite{held2021focus}, using RoBERTa \cite{liu2019roberta} as the primary encoder. A comparison between ACCI and several established baselines on the ECB+ dataset is reported in Table \ref{tab:table2}. On the test set, ACCI achieved $\text{B}^3$ F1 and $\text{CoNLL}$ F1 of 86.9\% and 88.4\%, respectively. Compared with the strongest baselines \cite{chen2023cross},\cite{nath2024okay}, which previously achieved the highest $\text{CoNLL}$ F1, our method delivers a notable improvement of 2.0\%. For the $\text{B}^3$ F1 metric, ACCI trails the best-reported score \cite{chen2023cross} by only 0.4\%, indicating highly competitive performance. Importantly, ACCI obtains the best scores across all competing methods on additional metrics including MUC, CEAFe, and LEA F1. This comparative evaluation underscores the substantial advantages of our approach. Although it marginally underperforms the optimal reported values on individual metrics, a holistic analysis across all evaluation criteria confirms that ACCI leads to marked improvements in the accuracy of cross-document event coreference resolution, thereby highlighting its efficacy in addressing the inherent challenges of the task.

To more systematically examine the performance differential between ACCI and the baseline models, Table \ref{tab:table3} presents an extensive comparison involving two representative backbone networks \cite{ahmed20232},\cite{held2021focus}. Furthermore, we investigate the performance of ACCI when adopting RoBERTa-large as the encoder. The results reveal that, relative to Held2021\cite{held2021focus} and the best-performing variants of Ahmed2023\cite{ahmed20232}, ACCI using RoBERTa-base achieves significant gains of 1.6\% and 1.0\% in $\text{CoNLL}$ F1. When further upgraded with RoBERTa-large, ACCI demonstrates even more pronounced improvements over Ahmed2023, boosting $\text{B}^3$ F1 and $\text{CoNLL}$ F1 by 1.9\% and 2.3\%, respectively.

Notably, even when employing RoBERTa-base, ACCI surpasses all Longformer-based counterparts, indicating that its superiority is not solely attributable to the choice of encoder. Instead, the effectiveness of ACCI arises from its architectural innovation and its tailored optimization strategy. These results also lend strong support to the necessity and effectiveness of incorporating causal intervention and counterfactual reasoning into ECR. 
 
\begin{table*}[htbp]
  \centering
  \caption{Performance comparison of models on the ECB+ dataset. - indicates that performance scores were not reported for the corresponding evaluation metric.}
  \begin{adjustbox}{width=\textwidth}
  \begin{tabular}{lccccccccccccc}
    \toprule
    \textbf{Model} & \multicolumn{3}{c}{\textbf{MUC}} & \multicolumn{3}{c}{\textbf{B$^3$}} & \multicolumn{3}{c}{\textbf{CEAF$_e$}} & \multicolumn{3}{c}{\textbf{LEA}} & \textbf{CoNLL} \\
    \cmidrule(r){2-14}
    & P & R & F1 & P & R & F1 & P & R & F1 & P & R & F1 & F1 \\
    \midrule
    \textbf{CDLM-based} & & & & & & & & & & & & & \\
    Caciularu2021 & 87.1 & 89.2 & 88.1 & 84.9 & 87.9 & 86.4 & 83.3 & 81.2 & 82.2 & 76.7 & 77.2 & 76.9 & 85.6 \\
    \midrule
    \textbf{Seq to Seq-based} & & & & & & & & & & & & & \\
    Cattan2021 & 81.9 & 85.1 & 83.5 & 82.7 & 82.1 & 82.4 & 78.9 & 75.2 & 77.0 & 72.0 & 68.8 & 70.4 & 81.0 \\
    \midrule
    \textbf{LLM-based} & & & & & & & & & & & & & \\
    Llama & 76.3 & 84.2 & 80.1 & 73.2 & 82.7 & 77.7 & 77.2 & 67.5 & 72.0 & - & - & - & 76.6 \\
    GPT-3.5-Turbo & 81.0 & 81.7 & 81.4 & 78.6 & 81.0 & 79.8 & 77.0 & 76.1 & 76.5 & - & - & - & 79.2 \\
    Nath2024 & 92.0 & 84.1 & 87.9 & 91.7 & 82.4 & 86.8 & 80.5 & 88.9 & \underline{84.5} & - & - & - & \underline{86.4} \\
    \midrule
    \textbf{Graph-based} & & & & & & & & & & & & & \\
    Chen2023 & 87.2 & 89.4 & \underline{88.3} & 86.4 & 88.3 & \textbf{87.3} & 84.0 & 83.2 & 83.6 & - & - & - & \underline{86.4} \\
    Gao2024 & 87.6 & 88.1 & 87.8 & 85.8 & 86.9 & 86.3 & 83.2 & 81.7 & 82.4 & - & - & - & 85.5 \\
    Chen2025 & 86.7 & 89.7 & 88.2 & 84.2 & 90.4 & \underline{87.2} & 85.3 & 81.7 & 83.5 & - & - & - & 86.3 \\
    Zhao2025 & 85.1 & 82.2 & 83.6 & 83.2 & 83.1 & 83.1 & 76.6 & 80.1 & 78.3 & - & - & - & 81.7 \\
    \midrule
    \textbf{Pairwise-based} & & & & & & & & & & & & & \\
    Barhom2019 & 78.1 & 84.0 & 80.9 & 76.8 & 86.1 & 81.2 & 79.6 & 73.3 & 76.3 & 64.6 & 72.3 & 68.3 & 79.5 \\
    Held2021 & 88.1 & 87.0 & 87.5 & 87.7 & 85.6 & 86.6 & 85.8 & 80.3 & 82.9 & 73.2 & 74.9 & 74.0 & 85.7 \\
    Yu2022 & 85.1 & 88.1 & 86.6 & 84.7 & 86.1 & 85.4 & 79.6 & 83.1 & 81.3 & - & - & - & 84.4 \\
    Ahmed2023 & 87.3 & 80.0 & 83.5 & 85.4 & 79.6 & 82.4 & 75.5 & 83.1 & 79.1 & 73.3 & 70.5 & 71.9 & 81.7 \\
    Ding2024 & 88.6 & 86.4 & 87.5 & 88.4 & 85.7 & 87.0 & 82.2 & 84.7 & 83.4 & 79.6 & 77.4 & 78.5 & 86.0 \\
    \rowcolor{gray!10}
    \textbf{ACCI} & 87.7 & 96.4 & \textbf{91.8} & 79.6 & 95.7 & 86.9 & 92.3 & 81.2 & \textbf{86.4} & 74.9 & 88.1 & 81.0 & \textbf{88.4} \\
    \bottomrule
  \end{tabular}
  \end{adjustbox}
  \label{tab:table2}
\end{table*}

\begin{table*}[htbp]
  \centering
  \caption{Comprehensive performance comparison between ACCI and two backbone networks on the ECB+ dataset. Base, Large, and Long denote models using RoBERTa-base, RoBERTa-large, and Longformer as encoders, respectively. Baseline results are obtained by reproducing the work.}
  \begin{adjustbox}{max width=\textwidth}
  \begin{tabular}{lccccccccccccc}
    \toprule
    \textbf{Model} & \multicolumn{3}{c}{\textbf{MUC}} & \multicolumn{3}{c}{\textbf{B$^3$}} & \multicolumn{3}{c}{\textbf{CEAF$_e$}} & \multicolumn{3}{c}{\textbf{LEA}} & \textbf{CoNLL} \\
    \cmidrule(r){2-14}
    & P & R & F1 & P & R & F1 & P & R & F1 & P & R & F1 & F1 \\
    \midrule
    \textbf{Held2021} & & & & & & & & & & & & & \\
    Baseline & 86.3 & 86.3 & 86.3 & 86.7 & 86.0 & 86.3 & 81.3 & 81.4 & 81.3 & 76.9 & 75.8 & 76.4 & 84.6 \\
    ACCI & 86.4 & 89.9 & 88.1 & 84.4 & 89.2 & 86.8 & 85.8 & 81.5 & 83.6 & 76.7 & 80.4 & 78.5 & 86.2 \\
    \midrule
    \textbf{Ahmed2023-LH} & & & & & & & & & & & & & \\
    Baseline-Roberta & 86.9 & 76.2 & 81.2 & 85.7 & 77.8 & 81.6 & 73.0 & 83.9 & 78.1 & 71.5 & 68.7 & 70.1 & 80.3 \\
    Baseline-Long & 87.3 & 80.0 & 83.5 & 85.4 & 79.6 & 82.4 & 75.5 & 83.1 & 79.1 & 73.3 & 70.5 & 71.9 & 81.7 \\
    ACCI-Base & 86.6 & 81.8 & 84.1 & 85.4 & 81.2 & 83.3 & 78.3 & 83.5 & 80.8 & 74.8 & 72.3 & 73.5 & 82.7 \\
    ACCI-Large & 86.5 & 83.2 & 84.8 & 85.2 & 81.9 & 83.5 & 78.8 & 82.4 & 80.5 & 75.1 & 72.8 & 73.9 & 82.9 \\
    \midrule
    \textbf{Ahmed2023-LH\_Oracle} & & & & & & & & & & & & & \\
    Baseline-Roberta & 87.6 & 89.8 & 88.7 & 80.2 & 90.7 & 85.1 & 85.1 & 82.5 & 83.8 & 72.2 & 83.3 & 77.3 & 85.9 \\
    Baseline-Long & 87.9 & 93.7 & 90.7 & 79.6 & 94.1 & 86.3 & 88.7 & 81.6 & 85.0 & 73.2 & 86.8 & 79.4 & 87.4 \\
    ACCI-Base & 87.7 & 96.4 & 91.8 & 79.6 & 95.7 & 86.9 & 92.3 & 81.2 & 86.4 & 74.9 & 88.1 & 81.0 & 88.4 \\
    ACCI-Large & 89.7 & 95.6 & 92.6 & 82.4 & 95.0 & 88.2 & 92.0 & 84.7 & 88.2 & 77.8 & 89.0 & 83.0 & 89.7 \\
    \bottomrule
  \end{tabular}
  \end{adjustbox}
  \label{tab:table3}
\end{table*}

\subsubsection{Results on the GVC Dataset}
To comprehensively evaluate the generalization capability of the proposed method, A detailed comparison of ACCI with existing advanced baseline methods on the GVC dataset is provided in Table \ref{tab:table4}. In the test set evaluation, the ACCI method achieved $\text{CoNLL}$ F1 and $\text{B}^3$ F1 of 85.2\% and 86.6\%, respectively. Compared to the baseline that reported the best $\text{CoNLL}$ F1 \cite{chen2025improving}, ACCI demonstrates a highly competitive performance, achieving a score within a small margin of only 0.6\% of their result. It is notable that on the $\text{B}^3$ F1 and $\text{LEA}$ F1 metrics, ACCI surpassed the best reported baseline results \cite{ahmed20232},\cite{ding2024rationale} by 0.8\% and 0.9\%, respectively. This indicates that this method has advantages in these key evaluation indicators. Although the $\text{CEAFe}$F1 of ACCI is lower than that of Chen2025 \cite{chen2025improving}, it remains higher than the scores of other compared models. This discrepancy may stem from ACCI's strong performance in capturing the causal relationships between local triggers and arguments, contrasted with certain limitations it exhibits in explicitly modeling document-level chain structures, thereby leading to a slight reduction in the $\text{CEAFe}$ (chain integrity matching) evaluation score. Nevertheless, when considering overall performance, ACCI maintains a highly competitive standing.

A comprehensive comparison between ACCI and two representative backbone architectures \cite{ahmed20232, held2021focus} on the GVC dataset is presented in Table~\ref{tab:table5}. The experimental results indicate that, when using RoBERTa-base as the encoder, ACCI achieves significant improvements of 3.1\% and 2.9\% in $\text{CoNLL}$ F1 over Held2021 \cite{held2021focus} and the strongest versions of Ahmed2023 \cite{ahmed20232}, respectively. While ACCI does not outperform all prior methods across every individual metric, the findings clearly demonstrate that our method substantially enhances the overall performance of prominent SOTA baselines, thereby providing strong evidence of its cross-dataset generalization capability.

\begin{table*}[htbp]
  \centering
  \caption{Performance comparison of models on the GVC dataset. - indicates that performance scores were not reported for the corresponding evaluation metric.}
  \begin{adjustbox}{width=\textwidth}
  \begin{tabular}{lccccccccccccc}
    \toprule
    \textbf{Model} & \multicolumn{3}{c}{\textbf{MUC}} & \multicolumn{3}{c}{\textbf{B$^3$}} & \multicolumn{3}{c}{\textbf{CEAF$_e$}} & \multicolumn{3}{c}{\textbf{LEA}} & \textbf{CoNLL} \\
    \cmidrule(r){2-14}
    & P & R & F1 & P & R & F1 & P & R & F1 & P & R & F1 & F1 \\
    \midrule
    \textbf{CDLM-based} & & & & & & & & & & & & & \\
    Caciularu2021 & 90.4 & 85.0 & 87.6 & 83.8 & 80.8 & 82.3 & 63.5 & 74.7 & 68.6 & - & - & - & 79.5 \\
    \midrule
    \textbf{LLM-based} & & & & & & & & & & & & & \\
    Llama & 84.3 & 93.9 & 88.8 & 38.1 & 89.5 & 53.4 & 54.9 & 28.9 & 37.9 & - & - & - & 60.0 \\
    GPT-3.5-Turbo & 81.9 & 88.6 & 85.1 & 35.4 & 82.6 & 49.6 & 41.1 & 27.1 & 32.7 & - & - & - & 55.8 \\
    Nath2024 & 94.2 & 91.6 & \textbf{92.9} & 82.1 & 86.7 & 84.3 & 68.1 & 75.8 & 71.7 & - & - & - & 83.0 \\
    \midrule
    \textbf{Graph-based} & & & & & & & & & & & & & \\
    Chen2025 & 91.4 & 92.6 & 92.0 & 81.4 & 89.4 & 85.2 & 81.9 & 78.7 & \textbf{80.3} & - & - & - & \textbf{85.8} \\
    \midrule
    \textbf{Pairwise-based} & & & & & & & & & & & & & \\
    Barhom2019 & - & - & - & 66.0 & 81.0 & 72.7 & - & - & - & - & - & - & - \\
    Held2021 & 91.2 & 91.8 & 91.5 & 83.8 & 82.2 & 83.0 & 77.9 & 75.5 & 76.7 & 82.3 & 79.0 & 80.6 & 83.7 \\
    Yu2022 & 88.5 & 92.9 & 90.6 & 80.3 & 82.1 & 81.2 & 71.8 & 79.5 & 75.5 & - & - & - & 82.4 \\
    Ahmed2023 & 91.1 & 84.0 & 87.4 & 76.4 & 79.0 & 77.7 & 52.5 & 69.6 & 59.9 & 63.9 & 74.1 & 68.6 & 75.0 \\
    Ding2024 & 92.1 & 90.4 & 91.3 & 86.8 & 84.8 & \underline{85.8} & 73.2 & 78.9 & 76.0 & - & - & - & 84.4 \\
    \rowcolor{gray!10}
    \textbf{ACCI} & 92.8 & 92.4 & \underline{92.6} & 85.0 & 88.3 & \textbf{86.6} & 75.6 & 77.2 & \underline{76.4} & 84.0 & 79.6 & \textbf{81.7} & \underline{85.2} \\
    \bottomrule
  \end{tabular}
  \end{adjustbox}
  \label{tab:table4}
\end{table*}

\begin{table*}[htbp]
  \centering
  \caption{Comprehensive performance comparison between ACCI and two backbone networks on the GVC dataset. Base, Large, and Long denote models using RoBERTa-base, RoBERTa-large, and Longformer as encoders, respectively. Baseline results are obtained by reproducing the work.}
  \begin{adjustbox}{max width=\textwidth}
  \begin{tabular}{lccccccccccccc}
    \toprule
    \textbf{Model} & \multicolumn{3}{c}{\textbf{MUC}} & \multicolumn{3}{c}{\textbf{B$^3$}} & \multicolumn{3}{c}{\textbf{CEAF$_e$}} & \multicolumn{3}{c}{\textbf{LEA}} & \textbf{CoNLL} \\
    \cmidrule(r){2-14}
     & P & R & F1 & P & R & F1 & P & R & F1 & P & R & F1 & F1 \\
    \midrule
    \textbf{Ahmed2023-LH} & & & & & & & & & & & & & \\
    Baseline-Roberta & 89.6 & 87.0 & 88.3 & 67.9 & 82.3 & 74.4 & 55.2 & 62.0 & 58.4 & 57.8 & 77.6 & 66.2 & 73.7 \\
    Baseline-Long    & 91.1 & 84.0 & 87.4 & 76.4 & 79.0 & 77.7 & 52.5 & 69.6 & 59.9 & 63.9 & 74.1 & 68.6 & 75.0 \\
    ACCI-Base        & 92.8 & 88.3 & 90.5 & 76.5 & 82.2 & 79.3 & 59.7 & 71.7 & 65.2 & 68.2 & 78.3 & 72.9 & 78.3 \\
    ACCI-Large       & 92.7 & 87.1 & 89.8 & 78.6 & 80.6 & 79.6 & 60.3 & 75.5 & 67.1 & 69.7 & 76.2 & 72.8 & 78.8 \\
    \midrule
    \textbf{Ahmed2023-LH\_Oracle} & & & & & & & & & & & & & \\
    Baseline-Roberta & 90.2 & 89.1 & 89.6 & 68.0 & 85.0 & 75.6 & 59.6 & 62.7 & 61.1 & 59.5 & 80.6 & 68.5 & 75.4 \\
    Baseline-Long    & 91.4 & 84.9 & 88.0 & 77.4 & 80.4 & 78.9 & 54.3 & 70.5 & 61.3 & 65.5 & 75.7 & 70.2 & 76.1 \\
    ACCI-Base        & 92.4 & 89.7 & 91.0 & 74.6 & 84.7 & 79.4 & 63.0 & 70.8 & 66.6 & 67.4 & 81.2 & 73.7 & 79.0 \\
    ACCI-Large       & 91.9 & 91.5 & 91.7 & 70.7 & 88.0 & 78.4 & 68.7 & 69.7 & 69.2 & 64.6 & 84.8 & 73.4 & 79.8 \\
    \midrule
    \textbf{Held2021} & & & & & & & & & & & & & \\
    Baseline         & 92.3 & 89.3 & 90.8 & 85.7 & 82.1 & 83.8 & 67.5 & 76.6 & 71.7 & 78.8 & 76.9 & 77.8 & 82.1 \\
    ACCI             & 92.8 & 92.4 & 92.6 & 85.0 & 88.3 & 86.6 & 75.6 & 77.2 & 76.4 & 84.0 & 79.6 & 81.7 & 85.2 \\
    \bottomrule
  \end{tabular}
  \end{adjustbox}
  \label{tab:table5}
\end{table*}

\subsection{Ablation Experiment}
\label{AE}
To evaluate the contribution of each model component to the performance of ACCI, we conducted ablation studies on both the ECB+ and GVC datasets. The examined components include \textbf{Trigger Bias Mitigation (TBM}) and \textbf{Context Argument Enhancement (CAE)}, where $\downarrow$ indicates a decrease in F1. Table \ref{tab:table6} reports the detailed results of this analysis.

Specifically, w/o TBM refers to the setting in which the contextual arguments of event mentions are retained while the event trigger words are removed. This corresponds to the counterfactual inquiry: \textit{"What would the model’s coreference decision be if it had not observed the trigger word associated with the event?"} Conversely, w/o CAE retains the trigger words while omitting the contextual arguments, representing the question: \textit{"What would the model’s prediction be if it had not seen the surrounding arguments related to the event?"} Finally, w/o TBM+CAE removes both components, i.e., without applying either trigger bias mitigation or context argument enhancement.

On the ECB+ dataset, we observe that removing TBM results in a performance drop of 0.6\% in $\text{CoNLL}$ F1 and 0.7\% in $\text{B}^3$ F1, with $\text{LEA}$ F1 showing the most substantial decline of 0.9\%. These findings underscore the importance of TBM in mitigating the adverse impact of lexical trigger bias on coreference decisions. Similarly, removing CAE leads to a 0.8\% decrease in $\text{CoNLL}$ F1 and a 0.5\% decrease in $\text{B}^3$ F1, with $\text{MUC}$ F1 and $\text{LEA}$ F1 decreasing by more than 1.0\%, indicating the essential role of contextual arguments in event understanding. These results further reveal that conventional ECR systems often overly depend on surface-level lexical cues from triggers, while underutilizing the informative context that arguments provide. When both TBM and CAE are removed, the performance degradation becomes even more pronounced. Overall, these results demonstrate that ACCI effectively enhances the model’s ability to capture the underlying causal relationships between event mentions and coreference decisions, by reducing overreliance on triggers and improving sensitivity to argument-level information.

\begin{table*}[htbp]
  \centering
  \setlength{\tabcolsep}{15pt} 
  \caption{Results of Ablation Studies}
   \begin{adjustbox}{width=\textwidth}
  \begin{tabular}{lccccc}
    \toprule
     & \textbf{MUC F1} & \textbf{B$^3$ F1} & \textbf{CEAFe F1} & \textbf{LEA F1} & \textbf{CoNLL F1} \\
     \cmidrule(r){2-6}
    \multicolumn{6}{c}{\textbf{ECB+}} \\
     \midrule
    ACCI                & 91.8 & 86.9 & 86.6 & 81.0 & 88.4 \\
    w/o TBM       & 91.6 \scriptsize{(0.2$\downarrow$)} & 86.3 \scriptsize{(0.6$\downarrow$)} & 85.8 \scriptsize{(0.6$\downarrow$)} & 80.2 \scriptsize{(0.8$\downarrow$)} & 87.9 \scriptsize{(0.5$\downarrow$)} \\
    w/o CAE       & 90.9 \scriptsize{(0.9$\downarrow$)} & 86.5 \scriptsize{(0.4$\downarrow$)} & 85.7 \scriptsize{(0.7$\downarrow$)} & 80.0 \scriptsize{(1.0$\downarrow$)} & 87.7 \scriptsize{(0.7$\downarrow$)} \\
    w/o TBM+CAE   & 91.3 \scriptsize{(0.5$\downarrow$)} & 85.3 \scriptsize{(1.6$\downarrow$)} & 85.0 \scriptsize{(1.4$\downarrow$)} & 79.3 \scriptsize{(1.7$\downarrow$)} & 87.2 \scriptsize{(1.2$\downarrow$)} \\
    \midrule
    \multicolumn{6}{c}{\textbf{GVC}} \\
    \midrule
    ACCI                & 91.0 & 79.4 & 66.6 & 73.7 & 79.0 \\
    w/o TBM       & 90.8 \scriptsize{(0.2$\downarrow$)} & 78.9 \scriptsize{(0.5$\downarrow$)} & 66.0 \scriptsize{(0.6$\downarrow$)} & 73.2 \scriptsize{(0.5$\downarrow$)} & 78.6 \scriptsize{(0.4$\downarrow$)} \\
    w/o CAE       & 90.4 \scriptsize{(0.6$\downarrow$)} & 79.2 \scriptsize{(0.2$\downarrow$)} & 65.2 \scriptsize{(1.4$\downarrow$)} & 72.8 \scriptsize{(0.9$\downarrow$)} & 78.3 \scriptsize{(0.7$\downarrow$)} \\
    w/o TBM+CAE   & 90.4 \scriptsize{(0.6$\downarrow$)} & 77.4 \scriptsize{(2.0$\downarrow$)} & 64.2 \scriptsize{(2.4$\downarrow$)} & 71.5 \scriptsize{(2.2$\downarrow$)} & 77.6 \scriptsize{(1.4$\downarrow$)} \\
    \bottomrule
  \end{tabular}
  \end{adjustbox}
  \label{tab:table6}
\end{table*}

\subsection{Hyperparameter analysis in the ACCI}
\begin{figure}
    \centering
    \includegraphics[width=1\columnwidth]{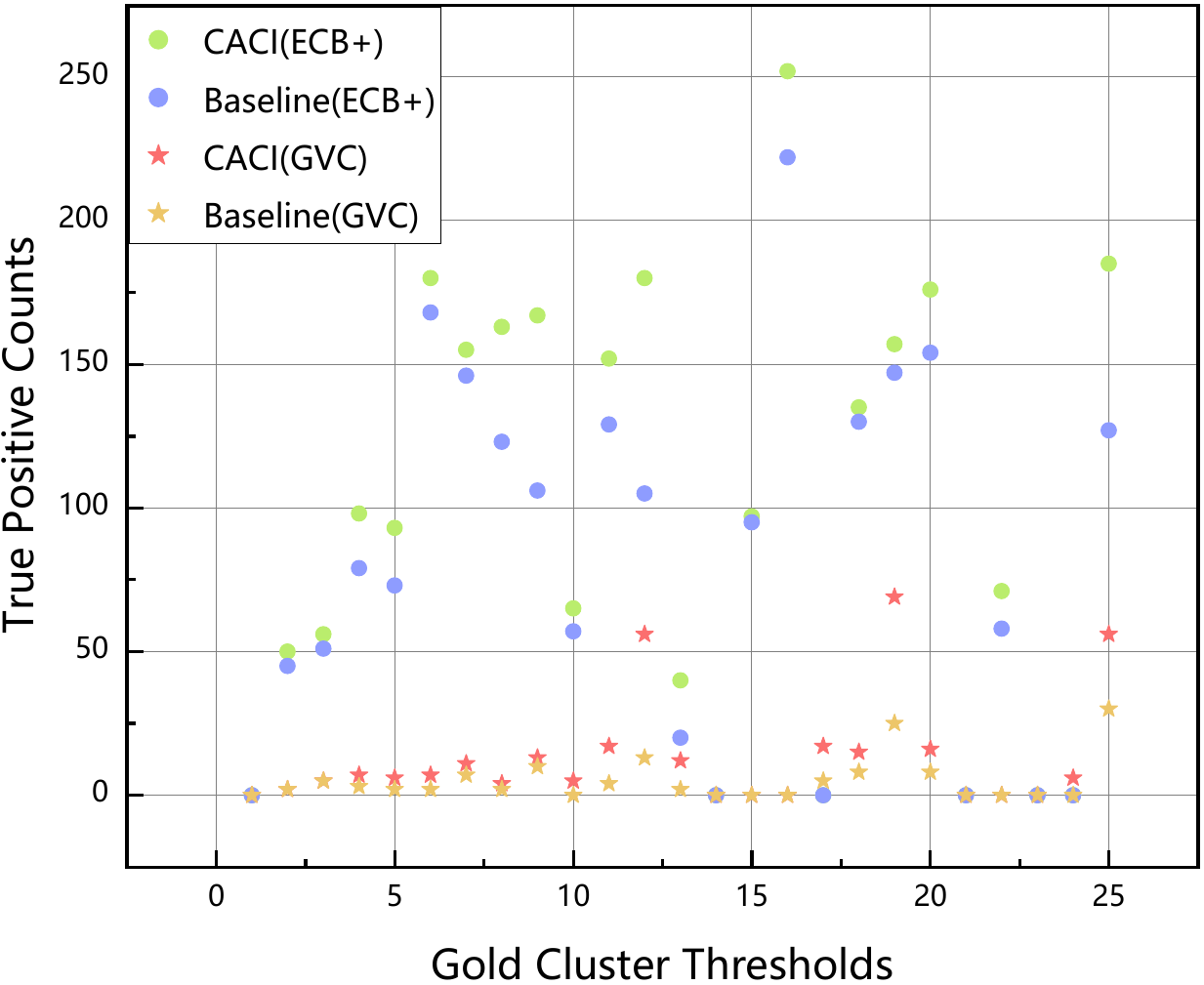}
    \caption{Distribution of correctly resolved mentions (True Positives) across cluster-size thresholds on the ECB+ and GVC datasets, comparing the ACCI model with baseline.}
    \label{fig:8}
\end{figure}

We systematically analyze the key hyperparameters of the ACCI model and perform sensitivity experiments to assess its performance across varying conditions. A detailed distribution of correctly resolved mentions (True Positives, TP) across varying cluster threshold levels for both the ACCI model and baseline methods on the ECB+ and GVC benchmark datasets is presented in Fig.~\ref{fig:8}. The quantitative results demonstrate that ACCI consistently outperforms the baselines on both corpora, with its advantage becoming increasingly prominent under higher clustering thresholds. As the clustering threshold increases, a clear upward trend in the number of true positives resolved by ACCI is observed. On the ECB+ dataset, the TP count peaks at a threshold of 16, while the performance gap between ACCI and the best-performing baseline reaches its maximum at threshold 25. On the GVC dataset, the largest disparity occurs at threshold 19. These findings suggest that higher threshold values impose stricter decision boundaries for event clustering, thereby serving as a robust measure of model resilience. The superior performance of ACCI under these challenging settings can be attributed to its ability to integrate rich contextual information, model argument structures more comprehensively, and suppress interference from misleading lexical cues. In contrast, baseline models tend to rely heavily on trigger word matching and superficial lexical similarity, which makes them more susceptible to incorrectly clustering non-coreferent but lexically similar event mentions when stricter thresholds are applied.

\begin{figure}
    \centering
    \includegraphics[width=1\columnwidth]{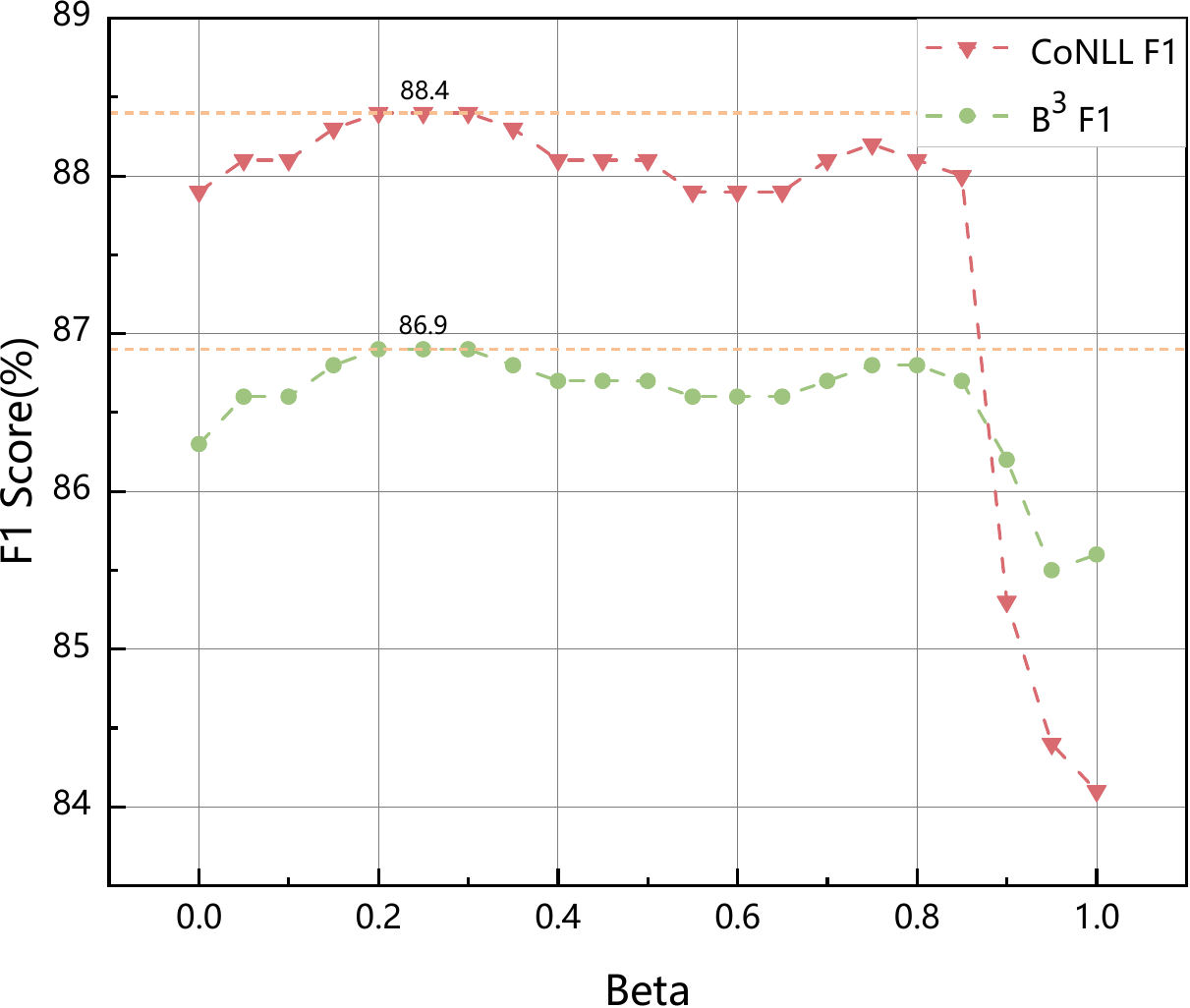}
    \caption{Effect of the debiasing coefficient $\beta$ on $\text{CoNLL}$ F1 and $\text{B}^3$ F1 on the ECB+ dataset. Optimal performance is observed when $\beta$ is set between 0.2 and 0.3.}
    \label{fig:9}
\end{figure}

To quantitatively assess the impact of the debiasing coefficient $\beta$, we varied its value from 0.0 to 1.0 in increments of 0.05 and evaluated the resulting $\text{CoNLL}$ F1 and $\text{B}^3$ F1 on the ECB+ dataset, and the results are illustrated in Fig. \ref{fig:9}. As $\beta$ increased from 0.0 (no debiasing) to the range of 0.2–0.3, model performance steadily improved, with the $\text{CoNLL}$ F1 reaching 88.4\% and the $\text{B}^3$ F1 reaching 86.9\%. These gains indicate that moderate debiasing effectively suppresses spurious correlations induced by trigger word matching while preserving valuable recall signals, thereby enhancing inference accuracy. As $\beta$ continued to rise beyond 0.3, both scores began to plateau and slightly declined when $\beta$ $\geq$ 0.5. Specifically, at $\beta$ = 0.5, the $\text{CoNLL}$ F1 and $\text{B}^3$ F1 stabilized around 88.1\% and 86.7\%, respectively. However, as $\beta$ approached 1.0—corresponding to near-total removal of trigger-based bias—the $\text{CoNLL}$ F1 dropped sharply to 84.1\%, and the $\text{B}^3$ F1 fell to 85.6\%. This suggests that although full debiasing can eliminate false associations, it also removes helpful semantic cues provided by trigger words, causing the model to underperform in event alignment.

These results highlight the need to balance debiasing strength and signal retention. When $\beta$ is set within the range of 0.2–0.3, the model can effectively suppress misleading correlations while retaining essential information from trigger words, achieving optimal performance on both evaluation metrics. This underscores the importance of carefully tuning the debiasing coefficient to maintain both robustness and accuracy in the debiasing framework.

\subsection{Comparison of Loss Functions}
\label{CLF}
\begin{figure}
    \centering
    \includegraphics[width=1\columnwidth]{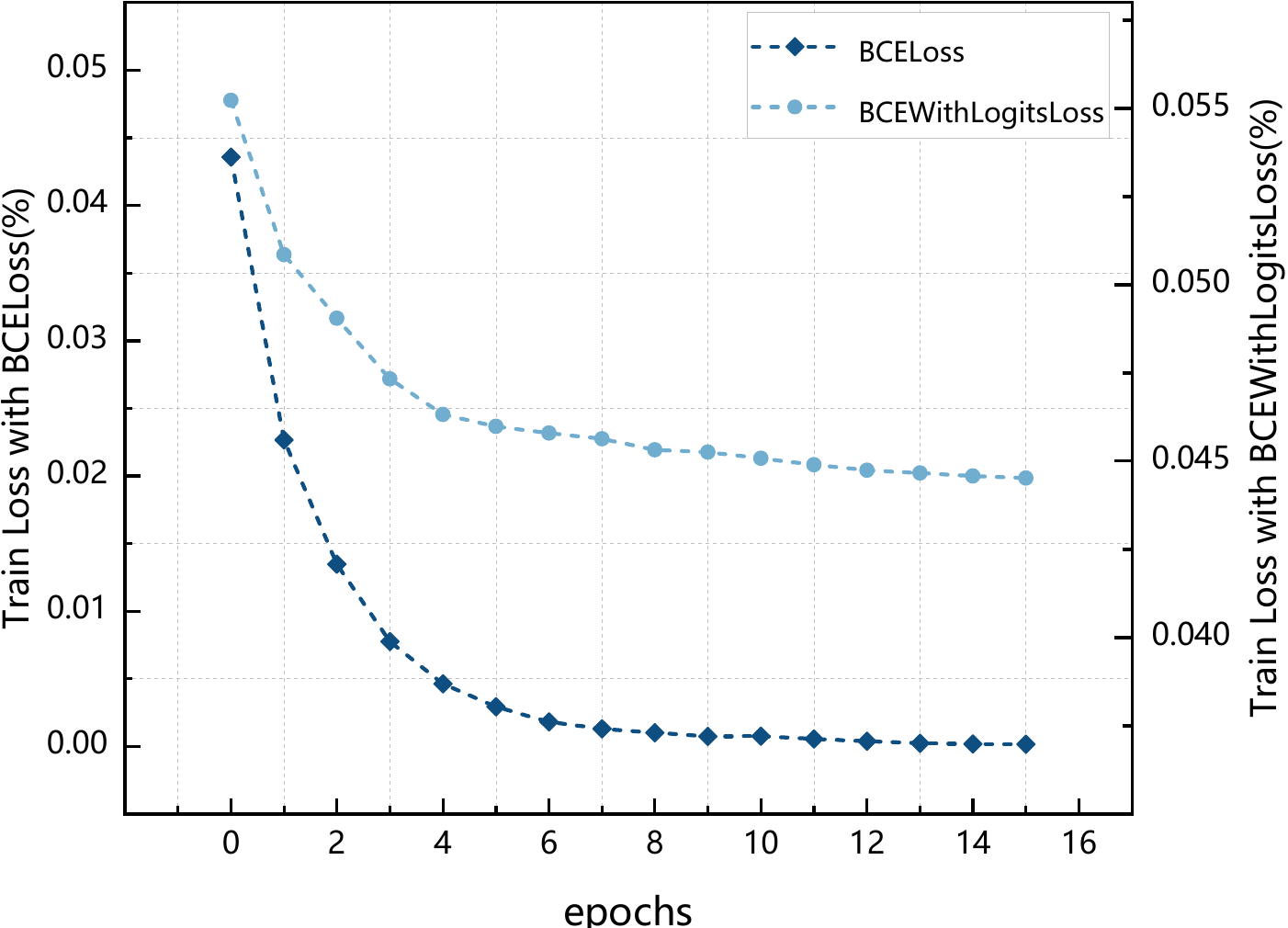}
    \caption{Training loss curves of BCELoss and BCEWithLogitsLoss under the ACCI framework.}
    \label{fig:10}
\end{figure}

In classification tasks, the selection of an appropriate loss function is a key determinant of model training efficacy. Under the proposed ACCI framework, we systematically compare two prevalent binary loss functions: Binary Cross-Entropy (BCELoss) and Binary Cross-Entropy with Logits (BCEWithLogitsLoss). The key difference between them lies in output normalization. BCELoss requires the output probabilities to be normalized via a sigmoid function, while BCEWithLogitsLoss operates directly on raw logits and incorporates the sigmoid operation internally to improve numerical stability. As illustrated in Fig. \ref{fig:10}, BCELoss consistently demonstrated superior performance throughout the training process in this task. It exhibited a lower initial loss, a faster rate of decrease, and ultimately converged to a smaller training loss with reduced overall fluctuations, indicating a more stable and efficient optimization trajectory. This phenomenon may be attributed to the explicit use of sigmoid activation by BCELoss, which enables the model to learn probability normalization behavior more directly during training, thereby facilitating the establishment of distinct decision boundaries between positive and negative samples. In contrast, while BCEWithLogitsLoss offers better numerical stability, its relatively smooth gradient characteristics might provide insufficient feedback for low-confidence samples during the initial training stages. This could limit the model's sensitivity to crucial discriminative signals, consequently impacting the overall convergence efficiency.

\subsection{Case Analysis}
\label{CA}
\begin{figure*}
    \centering
    \includegraphics[width=1.8\columnwidth]{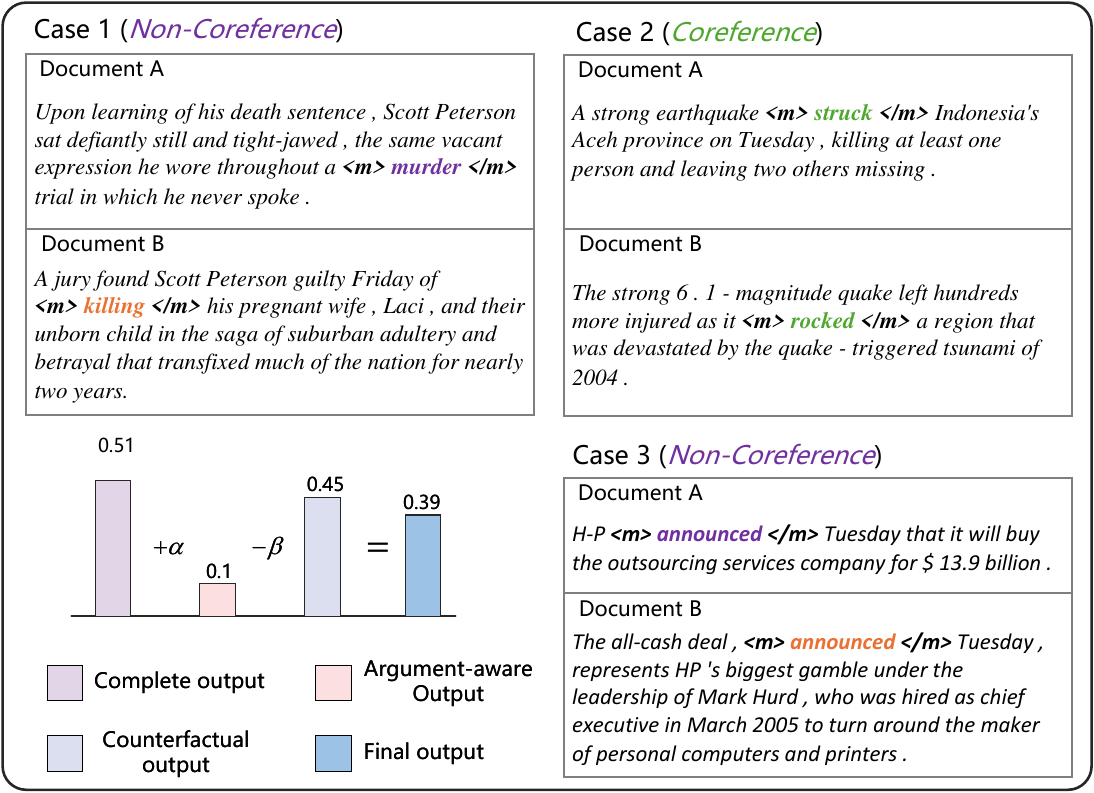}
    \caption{Three cross-document mention-pair cases from the ECB+ dataset, with event triggers marked between $<m>$ and $</m>$, “Non-Coreference” denoting non-coreferential pairs and “Coreference” denoting coreferential pairs; A bar chart is included for Case 1 to illustrate the progression of reasoning scores leading to the final prediction.}
    \label{fig:11}
\end{figure*}
Moreover, we construct representative case studies to assess the practical effectiveness of the proposed ACCI model. These cases are drawn from the ECB+ dataset and are selected based on their tendency to induce errors in the baseline model, while being correctly resolved by ACCI.

As illustrated in Fig. \ref{fig:11}, Case 1 presents two non-coreferent events extracted from Document A and Document B. While both refer to Scott Peterson, Document A employs the trigger word murder to portray his psychological state and demeanor during the sentencing phase of his trial. In contrast, Document B uses killing to directly reference the act of murdering his pregnant wife and unborn child. Although centered on the same individual, the semantic focus of the triggers clearly diverges: the former emphasizes courtroom proceedings, whereas the latter describes the criminal act itself. The baseline model mistakenly identifies the two as coreferent, reflecting its reliance on lexical similarity. In contrast, ACCI accurately distinguishes between the two, demonstrating a deeper understanding of contextual semantics. The accompanying bar chart visualizes the model’s internal reasoning process that led to its final decision.

In Case 2, the trigger struck is used to report the onset of an earthquake, while rocked captures the resulting tremor in the affected region. Both mentions refer to the same 6.1-magnitude earthquake in Aceh, Indonesia. Despite this, the baseline model incorrectly categorizes them as referring to different events, whereas ACCI successfully resolves them as coreferent.

In Case 3, both mentions share the trigger announced, but differ substantially in their underlying semantics. The first centers on HP’s \$13.9 billion acquisition of an outsourcing firm, focusing on the transactional details. The second, however, emphasizes the strategic implications of this cash-based deal. Here, the baseline model again fails by equating the two events, driven by superficial trigger word matching. This example underscores a key limitation of conventional ECR systems: when overly reliant on lexical similarity, they fail to capture fine-grained differences in argument structure and contextual nuance.

These cases highlight the necessity of ACCI. By mitigating spurious lexical correlations and enhancing sensitivity to argument-level cues, ACCI reduces confounding noise and improves causal inference between event mentions. This results in more robust, unbiased predictions and a significantly enhanced coreference resolution capability, particularly in semantically complex scenarios.

\subsection{Confounding Factor Impact Analysis}
\label{CFIA}
\begin{figure*}
    \centering
    \includegraphics[width=1.8\columnwidth]{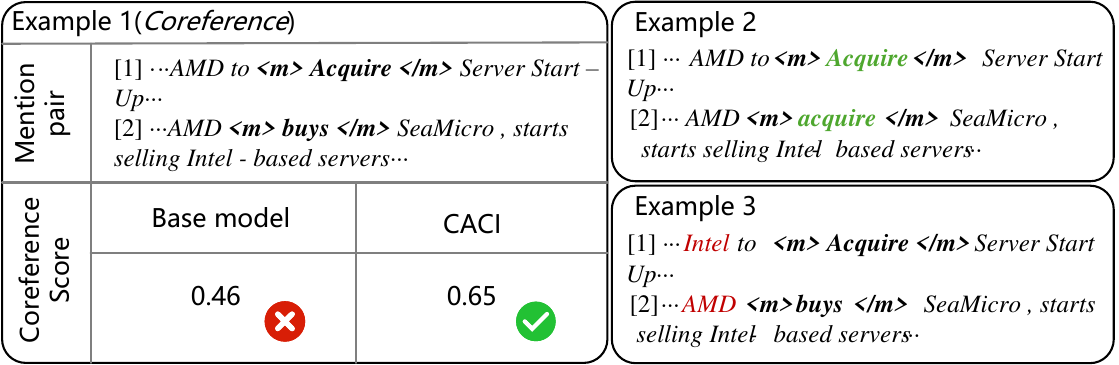}
    \caption{Case studies illustrating the impact of lexical trigger bias and argument sensitivity. ACCI shows improved robustness over the baseline by correctly handling semantically similar triggers, reducing sensitivity to superficial trigger matches, and detecting changes in event arguments.}
    \label{fig:12}
\end{figure*}
In this section, we provide illustrative examples to analyze the influence of potential confounders on the baseline ECR system. A key issue in the baseline model is its overreliance on lexical matching between trigger words in event mention pairs, which prevents it from capturing the underlying causal signals necessary for accurate coreference prediction. Within this context, we identify event triggers as latent confounding variables that bias the model’s decision-making process. In Example 1 (Fig. \ref{fig:12}), the two event mentions share semantically equivalent but lexically distinct trigger words: “acquire” and “buy”, both referring to the event in which AMD acquires the server company SeaMicro. However, influenced by the lexical discrepancy between the triggers, the baseline model assigns a coreference score of 0.46—below the classification threshold of 0.5, and thus incorrectly labels the pair as non-coreferential (gold label: 1.0). To further examine the model’s susceptibility to confounding effects, we conduct a minimal intervention in Example 2 by replacing the trigger “buys” with “acquire”. This slight lexical change significantly increases the model’s confidence, boosting the coreference score from 0.46 to 0.77. Such a sharp shift in prediction, despite identical contextual semantics, highlights the model’s high sensitivity to lexical forms of triggers—an indication of spurious correlations rather than true causal reasoning. In contrast, the proposed ACCI framework effectively attenuates the impact of such confounders, resulting in substantially more robust and context-aware predictions across semantically similar but lexically varied event mentions.

To assess the ACCI model’s sensitivity to event arguments embedded in textual mentions, we introduce a minimal intervention in Example 3, wherein the shared participant “AMD” in one mention is replaced with “Intel”. This change alters the ground-truth label from 1 to 0, indicating that the two mentions no longer refer to the same event. As anticipated, ACCI successfully detects the shift in event arguments, with the prediction score dropping markedly from 0.65 to –0.21. This observation provides compelling evidence that ACCI not only mitigates reliance on superficial trigger-word lexical patterns, but also strengthens the model’s capacity to discern semantically meaningful changes in argument structure. Such behavior reinforces the conclusion that causal intervention and counterfactual reasoning embedded within ACCI enable more reliable and context-sensitive predictions, thereby substantially improving the accuracy of the event coreference system.


\section{Conclusion}
\label{sec:conclusion}
This study introduces an argument-centric causal intervention frameworkk, named ACCI, to mitigate the problem of spurious correlations in CD-ECR. Specifically, a structural causal graph is constructed to identify the confounding effects of trigger words and the causal relationships in the embedded context semantics. Subsequently, backdoor adjustment and counterfactual intervention are employed to guide the model to focus on the causal interaction between event arguments and coreferential relationships, while effectively suppressing the spurious bias caused by the lexical matching of trigger words. Extensive experimental results validate the superior performance of ACCI compared to the competitive baseline. The empirical success of ACCI highlights the utility of causal inference and counterfactual reasoning as powerful tools for enhancing model reliability in ECR tasks. 

In the future, we aim to expand the capabilities of this framework by tackling more complex challenges, including multilingual and multimodal ECR. This expansion will enhance its universality and practical applicability. Furthermore, we will investigate methods to adapt ACCI to effectively process unstructured and noisy data sources, such as social media posts, news reports, or conversational corpora, which are often characterized by ambiguity and linguistic irregularities.

\section*{Ethical Consideration}
In this section, we review our paper for the following ethical considerations:
\begin{itemize}
  \item \textbf{Avoid using Generative AI and AI-assisted technologies during the writing process}. We hereby declare that we have never used any form of generative artificial intelligence technique or AI-assisted technology in the writing of this manuscript to ensure that our content maintains the highest degree of originality, accuracy, and ethical responsibility.
  \item \textbf{Avoid harm}. This manuscript focuses on the design of a deep learning framework for cross-document event coreference resolution in which participants were not harmed in any way.
  \item \textbf{Transparency}. We have provided the source code of the design framework and a link to the repository of the data used in the study, which we will upload as soon as the paper is accepted. To ensure sufficient transparency of our work, we have provided detailed experimental parameters as well as evaluation metrics and analyzed the impact of the involved parameters on the final performance in Section 5.
  \item \textbf{Public Good}. Our proposed ACCI framework is solely designed to output coreference information for a given pair of event mentions, which does not have any negative impact on social opinion.
  \item \textbf{Confidentiality and Data Security}. The datasets employed for model evaluation in this study have been extensively used in prior research, and therefore do not have any data security issues or confidentiality concerns.
\end{itemize}

In summary, we have determined that our manuscript does not present any potential ethical risks in terms of harm, public good, transparency, confidentiality, or data security.

\section*{Acknowledgements}
   This work is a research achievement supported by the "Tianshan Talent" Research Project of Xinjiang (No. 2022TSYCLJ0037), the National Natural Science Foundation of China (No. 62262065), the Science and Technology Program of Xinjiang (No. 2022B01008), the National Key R\&D Program of China (No. 2022ZD0115800), the National Science Foundation of China (No. 62341206), and the National Science Foundation of China (No. 62476233).

\bibliographystyle{unsrt}

\bibliography{cas-refs}

\begin{thebibliography}{10}

\bibitem{postma2018semeval}
Marten Postma, Filip Ilievski, and Piek Vossen.
\newblock Semeval-2018 task 5: Counting events and participants in the long tail.
\newblock In {\em Proceedings of The 12th International Workshop on Semantic Evaluation}, pages 70--80, 2018.

\bibitem{humphreys1997event}
Kevin Humphreys, Robert Gaizauskas, and Saliha Azzam.
\newblock Event coreference for information extraction.
\newblock In {\em Operational Factors in Practical, Robust Anaphora Resolution for Unrestricted Texts}, 1997.

\bibitem{yang2018hotpotqa}
Zhilin Yang, Peng Qi, Saizheng Zhang, Yoshua Bengio, William~W Cohen, Ruslan Salakhutdinov, and Christopher~D Manning.
\newblock Hotpotqa: A dataset for diverse, explainable multi-hop question answering.
\newblock {\em arXiv preprint arXiv:1809.09600}, 2018.

\bibitem{barhom2019revisiting}
Shany Barhom, Vered Shwartz, Alon Eirew, Michael Bugert, Nils Reimers, and Ido Dagan.
\newblock Revisiting joint modeling of cross-document entity and event coreference resolution.
\newblock In {\em Proceedings of the 57th Annual Meeting of the Association for Computational Linguistics}, pages 4179--4189, 2019.

\bibitem{ahmed20232}
Shafiuddin~Rehan Ahmed, Abhijnan Nath, James~H Martin, and Nikhil Krishnaswamy.
\newblock 2* n is better than n2: Decomposing event coreference resolution into two tractable problems.
\newblock In {\em Findings of the Association for Computational Linguistics: ACL 2023}, pages 1569--1583, 2023.

\bibitem{Yu0R22}
Xiaodong Yu, Wenpeng Yin, and Dan Roth.
\newblock Pairwise representation learning for event coreference.
\newblock In Vivi Nastase, Ellie Pavlick, Mohammad~Taher Pilehvar, Jos{\'{e}} Camacho{-}Collados, and Alessandro Raganato, editors, {\em Proceedings of the 11th Joint Conference on Lexical and Computational Semantics, *SEM@NAACL-HLT 2022, Seattle, WA, USA, July 14-15, 2022}, pages 69--78. Association for Computational Linguistics, 2022.

\bibitem{ding2024rationale}
Bowen Ding, Qingkai Min, Shengkun Ma, Yingjie Li, Linyi Yang, and Yue Zhang.
\newblock A rationale-centric counterfactual data augmentation method for cross-document event coreference resolution.
\newblock In {\em Proceedings of the 2024 Conference of the North American Chapter of the Association for Computational Linguistics: Human Language Technologies (Volume 1: Long Papers)}, pages 1112--1140, 2024.

\bibitem{ravi2023happens}
Sahithya Ravi, Chris Tanner, Raymond Ng, and Vered Shwartz.
\newblock What happens before and after: Multi-event commonsense in event coreference resolution.
\newblock In {\em Proceedings of the 17th Conference of the European Chapter of the Association for Computational Linguistics}, pages 1708--1724, 2023.

\bibitem{pearl2016causal}
Judea Pearl, Madelyn Glymour, and Nicholas~P Jewell.
\newblock {\em Causal inference in statistics: A primer}.
\newblock John Wiley \& Sons, 2016.

\bibitem{bagga1999cross}
Amit Bagga and Breck Baldwin.
\newblock Cross-document event coreference: Annotations, experiments, and observations.
\newblock In {\em Coreference and Its Applications}, 1999.

\bibitem{lu2017joint}
Jing Lu and Vincent Ng.
\newblock Joint learning for event coreference resolution.
\newblock In {\em Proceedings of the 55th Annual Meeting of the Association for Computational Linguistics (Volume 1: Long Papers)}, pages 90--101, 2017.

\bibitem{fang2018employing}
Jie Fang, Peifeng Li, and Guodong Zhou.
\newblock Employing multiple decomposable attention networks to resolve event coreference.
\newblock In {\em CCF International Conference on Natural Language Processing and Chinese Computing}, pages 246--256. Springer, 2018.

\bibitem{xu2023corefprompt}
Sheng Xu, Peifeng Li, and Qiaoming Zhu.
\newblock Corefprompt: Prompt-based event coreference resolution by measuring event type and argument compatibilities.
\newblock In {\em Proceedings of the 2023 Conference on Empirical Methods in Natural Language Processing}, pages 15440--15452, 2023.

\bibitem{yao2023learning}
Yao Yao, Zuchao Li, and Hai Zhao.
\newblock Learning event-aware measures for event coreference resolution.
\newblock In {\em Findings of the Association for Computational Linguistics: ACL 2023}, pages 13542--13556, 2023.

\bibitem{caciularu2021cdlm}
Avi Caciularu, Arman Cohan, Iz~Beltagy, Matthew~E Peters, Arie Cattan, and Ido Dagan.
\newblock Cdlm: Cross-document language modeling.
\newblock In {\em Findings of the Association for Computational Linguistics: EMNLP 2021}, pages 2648--2662, 2021.

\bibitem{cattan2021cross}
Arie Cattan, Alon Eirew, Gabriel Stanovsky, Mandar Joshi, and Ido Dagan.
\newblock Cross-document coreference resolution over predicted mentions.
\newblock In {\em Findings of the Association for Computational Linguistics: ACL-IJCNLP 2021}, pages 5100--5107, 2021.

\bibitem{lu2022end}
Yaojie Lu, Hongyu Lin, Jialong Tang, Xianpei Han, and Le~Sun.
\newblock End-to-end neural event coreference resolution.
\newblock {\em Artificial Intelligence}, 303:103632, 2022.

\bibitem{held2021focus}
William Held, Dan Iter, and Dan Jurafsky.
\newblock Focus on what matters: Applying discourse coherence theory to cross document coreference.
\newblock In {\em Proceedings of the 2021 Conference on Empirical Methods in Natural Language Processing}, pages 1406--1417, 2021.

\bibitem{chen2023cross}
Xinyu Chen, Sheng Xu, Peifeng Li, and Qiaoming Zhu.
\newblock Cross-document event coreference resolution on discourse structure.
\newblock In {\em Proceedings of the 2023 Conference on Empirical Methods in Natural Language Processing}, pages 4833--4843, 2023.

\bibitem{gao2024enhancing}
Qiang Gao, Bobo Li, Zixiang Meng, Yunlong Li, Jun Zhou, Fei Li, Chong Teng, and Donghong Ji.
\newblock Enhancing cross-document event coreference resolution by discourse structure and semantic information.
\newblock In {\em Proceedings of the 2024 Joint International Conference on Computational Linguistics, Language Resources and Evaluation (LREC-COLING 2024)}, pages 5907--5921, 2024.

\bibitem{chen2025improving}
Xinyu Chen, Peifeng Li, and Qiaoming Zhu.
\newblock Improving cross-document event coreference resolution by discourse coherence and structure.
\newblock {\em Information Processing \& Management}, 62(4):104085, 2025.

\bibitem{zhao2025hypergraph}
Wenbin Zhao, Yuhang Zhang, Di~Wu, Feng Wu, and Neha Jain.
\newblock Hypergraph convolutional networks with multi-ordering relations for cross-document event coreference resolution.
\newblock {\em Information Fusion}, 115:102769, 2025.

\bibitem{ahmed2024linear}
Shafiuddin~Rehan Ahmed, George~Arthur Baker, Evi Judge, Michael Reagan, Kristin Wright-Bettner, Martha Palmer, and James~H Martin.
\newblock Linear cross-document event coreference resolution with x-amr.
\newblock In {\em Proceedings of the 2024 Joint International Conference on Computational Linguistics, Language Resources and Evaluation (LREC-COLING 2024)}, pages 10517--10529, 2024.

\bibitem{nath2024okay}
Abhijnan Nath, Shadi~Manafi Avari, Avyakta Chelle, and Nikhil Krishnaswamy.
\newblock Okay, let’s do this! modeling event coreference with generated rationales and knowledge distillation.
\newblock In {\em Proceedings of the 2024 Conference of the North American Chapter of the Association for Computational Linguistics: Human Language Technologies (Volume 1: Long Papers)}, pages 3931--3946, 2024.

\bibitem{neuberg2003causality}
Leland~Gerson Neuberg.
\newblock Causality: models, reasoning, and inference, by judea pearl, cambridge university press, 2000.
\newblock {\em Econometric Theory}, 19(4):675--685, 2003.

\bibitem{lin2022causal}
Xiangru Lin, Ziyi Wu, Guanqi Chen, Guanbin Li, and Yizhou Yu.
\newblock A causal debiasing framework for unsupervised salient object detection.
\newblock In {\em Proceedings of the AAAI Conference on Artificial Intelligence}, volume~36, pages 1610--1619, 2022.

\bibitem{yang2021causal}
Xu~Yang, Hanwang Zhang, Guojun Qi, and Jianfei Cai.
\newblock Causal attention for vision-language tasks.
\newblock In {\em Proceedings of the IEEE/CVF conference on computer vision and pattern recognition}, pages 9847--9857, 2021.

\bibitem{qian2021counterfactual}
Chen Qian, Fuli Feng, Lijie Wen, Chunping Ma, and Pengjun Xie.
\newblock Counterfactual inference for text classification debiasing.
\newblock In {\em Proceedings of the 59th Annual Meeting of the Association for Computational Linguistics and the 11th International Joint Conference on Natural Language Processing (Volume 1: Long Papers)}, pages 5434--5445, 2021.

\bibitem{tang2020unbiased}
Kaihua Tang, Yulei Niu, Jianqiang Huang, Jiaxin Shi, and Hanwang Zhang.
\newblock Unbiased scene graph generation from biased training.
\newblock In {\em Proceedings of the IEEE/CVF conference on computer vision and pattern recognition}, pages 3716--3725, 2020.

\bibitem{pearl2009causal}
Judea Pearl.
\newblock Causal inference in statistics: An overview.
\newblock 2009.

\bibitem{pearl2009causality}
Judea Pearl.
\newblock {\em Causality}.
\newblock Cambridge university press, 2009.

\bibitem{zhang2021causal}
Yang Zhang, Fuli Feng, Xiangnan He, Tianxin Wei, Chonggang Song, Guohui Ling, and Yongdong Zhang.
\newblock Causal intervention for leveraging popularity bias in recommendation.
\newblock In {\em Proceedings of the 44th international ACM SIGIR conference on research and development in information retrieval}, pages 11--20, 2021.

\bibitem{wang2021clicks}
Wenjie Wang, Fuli Feng, Xiangnan He, Hanwang Zhang, and Tat-Seng Chua.
\newblock Clicks can be cheating: Counterfactual recommendation for mitigating clickbait issue.
\newblock In {\em Proceedings of the 44th International ACM SIGIR Conference on Research and Development in Information Retrieval}, pages 1288--1297, 2021.

\bibitem{zhang2021biasing}
Wenkai Zhang, Hongyu Lin, Xianpei Han, and Le~Sun.
\newblock De-biasing distantly supervised named entity recognition via causal intervention.
\newblock In {\em Proceedings of the 59th Annual Meeting of the Association for Computational Linguistics and the 11th International Joint Conference on Natural Language Processing (Volume 1: Long Papers)}, pages 4803--4813, 2021.

\bibitem{huang2020counterfactually}
William Huang, Haokun Liu, and Samuel Bowman.
\newblock Counterfactually-augmented snli training data does not yield better generalization than unaugmented data.
\newblock In {\em Proceedings of the First Workshop on Insights from Negative Results in NLP}, pages 82--87, 2020.

\bibitem{yang2021deconfounded}
Xu~Yang, Hanwang Zhang, and Jianfei Cai.
\newblock Deconfounded image captioning: A causal retrospect.
\newblock {\em IEEE Transactions on Pattern Analysis and Machine Intelligence}, 45(11):12996--13010, 2021.

\bibitem{yang2023context}
Dingkang Yang, Zhaoyu Chen, Yuzheng Wang, Shunli Wang, Mingcheng Li, Siao Liu, Xiao Zhao, Shuai Huang, Zhiyan Dong, Peng Zhai, et~al.
\newblock Context de-confounded emotion recognition.
\newblock In {\em Proceedings of the IEEE/CVF Conference on Computer Vision and Pattern Recognition}, pages 19005--19015, 2023.

\bibitem{zhang2024if}
Letian Zhang, Xiaotong Zhai, Zhongkai Zhao, Yongshuo Zong, Xin Wen, and Bingchen Zhao.
\newblock What if the tv was off? examining counterfactual reasoning abilities of multi-modal language models.
\newblock In {\em Proceedings of the IEEE/CVF Conference on Computer Vision and Pattern Recognition}, pages 21853--21862, 2024.

\bibitem{chen2023causal}
Ziwei Chen, Linmei Hu, Weixin Li, Yingxia Shao, and Liqiang Nie.
\newblock Causal intervention and counterfactual reasoning for multi-modal fake news detection.
\newblock In {\em Proceedings of the 61st Annual Meeting of the Association for Computational Linguistics (Volume 1: Long Papers)}, pages 627--638, 2023.

\bibitem{mu2023enhancing}
Feiteng Mu and Wenjie Li.
\newblock Enhancing event causality identification with counterfactual reasoning.
\newblock In {\em Proceedings of the 61st Annual Meeting of the Association for Computational Linguistics (Volume 2: Short Papers)}, pages 967--975, 2023.

\bibitem{joyce2003bayes}
James Joyce.
\newblock Bayes’ theorem.
\newblock 2003.

\bibitem{pearl2022probabilities}
Judea Pearl.
\newblock Probabilities of causation: three counterfactual interpretations and their identification.
\newblock In {\em Probabilistic and causal inference: the works of Judea Pearl}, pages 317--372. 2022.

\bibitem{cybulska2015bag}
Agata Cybulska and Piek Vossen.
\newblock " bag of events" approach to event coreference resolution. supervised classification of event templates.
\newblock {\em Int. J. Comput. Linguistics Appl.}, 6(2):11--27, 2015.

\bibitem{cybulska2014using}
Agata Cybulska and Piek Vossen.
\newblock Using a sledgehammer to crack a nut? lexical diversity and event coreference resolution.
\newblock In {\em Proceedings of the Ninth International Conference on Language Resources and Evaluation (LREC'14)}, pages 4545--4552, 2014.

\bibitem{vossen2018don}
Piek Vossen, Filip Ilievski, Marten Postma, and Roxane Segers.
\newblock Don’t annotate, but validate: A data-to-text method for capturing event data.
\newblock In {\em Proceedings of the Eleventh International Conference on Language Resources and Evaluation (LREC 2018)}, 2018.

\bibitem{bugert2021generalizing}
Michael Bugert, Nils Reimers, and Iryna Gurevych.
\newblock Generalizing cross-document event coreference resolution across multiple corpora.
\newblock {\em Computational Linguistics}, 47(3):575--614, 2021.

\bibitem{bagga1998entity}
Amit Bagga and Breck Baldwin.
\newblock Entity-based cross-document coreferencing using the vector space model.
\newblock In {\em COLING 1998 Volume 1: The 17th international conference on computational linguistics}, 1998.

\bibitem{moosavi2016coreference}
Nafise~Sadat Moosavi and Michael Strube.
\newblock Which coreference evaluation metric do you trust? a proposal for a link-based entity aware metric.
\newblock In {\em Proceedings of the 54th annual meeting of the association for computational linguistics}, volume~1, pages 632--642. Association for Computational Linguistics, 2016.

\bibitem{vilain1995model}
Marc Vilain, John~D Burger, John Aberdeen, Dennis Connolly, and Lynette Hirschman.
\newblock A model-theoretic coreference scoring scheme.
\newblock In {\em Sixth Message Understanding Conference (MUC-6): Proceedings of a Conference Held in Columbia, Maryland, November 6-8, 1995}, 1995.

\bibitem{luo2005coreference}
Xiaoqiang Luo.
\newblock On coreference resolution performance metrics.
\newblock In {\em Proceedings of human language technology conference and conference on empirical methods in natural language processing}, pages 25--32, 2005.

\bibitem{imambi2021pytorch}
Sagar Imambi, Kolla~Bhanu Prakash, and GR~Kanagachidambaresan.
\newblock Pytorch.
\newblock {\em Programming with TensorFlow: solution for edge computing applications}, pages 87--104, 2021.

\bibitem{LoshchilovH19}
Ilya Loshchilov and Frank Hutter.
\newblock Decoupled weight decay regularization.
\newblock In {\em 7th International Conference on Learning Representations, {ICLR} 2019, New Orleans, LA, USA, May 6-9, 2019}. OpenReview.net, 2019.

\bibitem{liu2019roberta}
Yinhan Liu, Myle Ott, Naman Goyal, Jingfei Du, Mandar Joshi, Danqi Chen, Omer Levy, Mike Lewis, Luke Zettlemoyer, and Veselin Stoyanov.
\newblock Roberta: A robustly optimized bert pretraining approach.
\newblock {\em arXiv preprint arXiv:1907.11692}, 2019.

\end{thebibliography}
\end{sloppypar}
\end{document}